%% file: main.tex
\documentclass[]{beingbeyond}
\usepackage{enumitem}
\usepackage[toc,page,header]{appendix}

\usepackage[utf8]{inputenc} 
\usepackage[T1]{fontenc}    
\usepackage{hyperref}       
\usepackage{url}            
\usepackage{array}          
\usepackage{booktabs}       
\usepackage{amsfonts}       
\usepackage{nicefrac}       
\usepackage{microtype}      
\usepackage{xcolor}         
\usepackage{xspace}
\usepackage{bm}
\usepackage{bbm}
\usepackage{tabularx}
\usepackage{amssymb}
\usepackage{enumitem}
\usepackage{amsmath}
\usepackage{mathtools}
\usepackage{amsthm}
\usepackage{multirow}
\usepackage{makecell}
\usepackage{color}
\usepackage{colortbl}
\usepackage{adjustbox}
\usepackage{caption}
\usepackage{graphicx}
\usepackage{wrapfig}

\definecolor{BlockC}{gray}{0.98}  
\definecolor{BlockA}{RGB}{191,211,230}
\definecolor{BlockB}{RGB}{199,233,192} 
\newcommand{\etal}{\emph{et al.}}

\title{Being-M0.5: A Real-Time Controllable Vision-Language-Motion Model}

\author{{\bfseries Bin Cao$^{1,2,3*}$ \quad Sipeng Zheng$^{7*}$ \quad Ye Wang$^{5}$ \quad Lujie Xia$^{4}$ \quad Qianshan Wei$^{6}$ \\ \quad Qin Jin$^{5}$ \quad Jing Liu$^{1,2}$ \quad Zongqing Lu$^{4,7\dagger}$}}

\affiliation{{$^{1}$CASIA \quad $^{2}$UCAS \quad $^{3}$BAAI \quad $^{4}$PKU \quad $^{5}$RUC \quad $^{6}$SEU \quad $^{7}$BeingBeyond}}

\abstract{
\input{sec/0_abstract}    
}

\checkdata[Project Page]{\url{https://beingbeyond.github.io/Being-M0.5}}
\checkdata[Date]{August 11, 2025}

\begingroup
 
\footnotetext[1]{These authors contributed equally to this work.}
\footnotetext[2]{Correspondence to Zongqing Lu $<$lu@beingbeyond.com$>$.}
\endgroup

\begin{document}

\maketitle

\input{sec/1_intro}
\input{sec/2_relatedwork}
\input{sec/3_method}

\input{sec/4_dataset}
\input{sec/5_experiments}
\input{sec/6_conclusion}

\clearpage

\bibliographystyle{unsrt}
\bibliography{ref}

\clearpage

\beginappendix

\input{sec/7_appendix}

\clearpage

\end{document}

%% file: sec/0_abstract.tex
Human motion generation has emerged as a critical technology with transformative potential for real-world applications. 
However, existing vision-language-motion models (VLMMs) face significant limitations that hinder their practical deployment.
We identify controllability as a main bottleneck, manifesting in five key aspects: inadequate response to diverse human commands, limited pose initialization capabilities, poor performance on long-term sequences, insufficient handling of unseen scenarios, and lack of fine-grained control over individual body parts.
To overcome these limitations, we present \texttt{\textbf{Being-M0.5}}, the first real-time, controllable VLMM that achieves state-of-the-art performance across multiple motion generation tasks. 
Our approach is built upon \texttt{\textbf{HuMo100M}}, the largest and most comprehensive human motion dataset to date, comprising over 5 million self-collected motion sequences, 100 million multi-task instructional instances, and detailed part-level annotations that address a critical gap in existing datasets.
We introduce a novel part-aware residual quantization technique for motion tokenization that enables precise, granular control over individual body parts during generation. 
Extensive experimental validation demonstrates  \texttt{\textbf{Being-M0.5}}'s superior performance across diverse motion benchmarks, while comprehensive efficiency analysis confirms its real-time capabilities.
Our contributions include design insights and detailed computational analysis to guide future development of practical motion generators. 
We believe that \texttt{\textbf{HuMo100M}} and \texttt{\textbf{Being-M0.5}} represent significant advances that will accelerate the adoption of motion generation technologies in real-world applications.

%% file: sec/1_intro.tex
\section{Introduction}
\label{sec:intro}

\begin{figure}[t]
\centering
     \includegraphics[width=1.0\columnwidth]{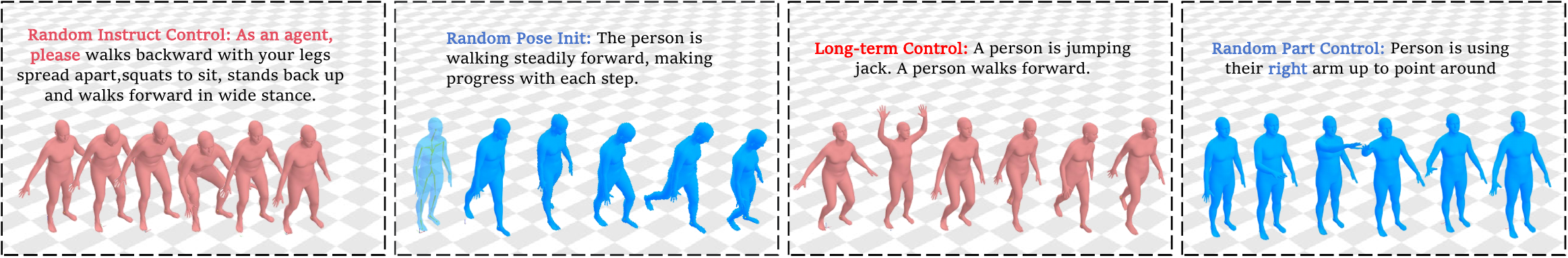}
 \caption{Leveraging our million-scale dataset \texttt{\textbf{HuMo100M}}, we present \texttt{\textbf{Being-M0.5}}, the first real-time, controllable vision-language-motion model (VLMM) that achieves both high performance and practical efficiency. \texttt{\textbf{Being-M0.5}} enables comprehensive controllability through five key capabilities: diverse natural language instruction following, flexible pose initialization, long-term motion sequence generation, handling of unseen motion patterns, and precise part-aware motion control.}
 \label{fig:intro}
 \end{figure}

Motion generation has garnered increasing attention due to its transformative applications in video games, film production, and humanoid robotics. 
However, current human motion generators~\cite{guo2024momask,jiang2023motiongpt} face significant challenges in achieving real-time inference speeds and comprehensive controllability, limiting their practical deployment.
We define controllability as the ability to handle diverse user commands, accommodate arbitrary initial poses, generate long-term or novel motion sequences, and enable precise part-level control. 
While methods trained on domain-specific datasets~\cite{guo2022generating,lin2024motion} perform well within fixed scenarios (e.g., text-to-motion), they exhibit poor generalization beyond these constraints.
Inspired by the success of large vision-language models (VLMs)~\cite{li2023blip,liu2023improvedllava}, recent efforts have developed vision-language-motion models (VLMMs) through multi-modal and multi-task training paradigms.
These models~\cite{ji2018large} have demonstrated enhanced motion generation capabilities, with some incorporating visual cues~\cite{luo2024m} for improved motion understanding.
Despite these advances, achieving comprehensive controllability remains an open challenge, motivating the present work.

Large-scale multimodal data is fundamental to unlocking the potential of VLMs. 
However, motion generation is severely constrained by the scarcity of high-quality motion datasets. 
While ecent efforts~\cite{wang2024quo} have explored extracting motions from web videos to construct larger datasets, most approaches fail to effectively utilize the collected data beyond simple scaling. 
To address this, we introduce \texttt{\textbf{HuMo100M}}, the largest motion generation dataset to date, comprising over 5 million motion sequences and 100 million instructional instances across diverse tasks.
Our dataset provides three key innovations over existing datasets:
\textbf{(1) Part-level descriptions} that offer fine-grained supervision for part-specific control, enabling precise alignment with part-level motions while filtering low-quality, occluded, or blurred segments to enhance data reliability.
\textbf{(2) Long-term motion sequences} generated through a novel motion concatenation method that combines individual motions into continuous, spatiotemporally consistent sequences, enabling VLMMs to produce realistic extended motions beyond short-term clips.
\textbf{(3) Text-aligned visual clips} that, unlike prior approaches, leverage visual cues as particularly beneficial for Internet-collected motions, allowing VLMMs to learn through weak supervision via visual-textual context alignment even when motion data quality is suboptimal.

Building upon the proposed dataset, we introduce \texttt{\textbf{Being-M0.5}}, the first real-time, controllable VLMM achieving state-of-the-art performance across diverse motion benchmarks. 
Leveraging \texttt{\textbf{HuMo100M}}’s million-scale instructional data and multi-modal inputs, our model explores critical design choices for practical, time-efficient motion generation (e.g., motion decoding strategies), a topic that has received limited attention in prior work.
We emphasize the importance of part-level control in human motion activities, which presents significant challenges for existing methods due to their reliance on single code embeddings representing the entire body and the absence of part-specific annotations.
Inspired by residual vector quantization (RQ) techniques~\cite{lee2022autoregressive, guo2024momask}, we propose part-aware residual quantization (PRQ) for motion tokenization. 
Like standard RQ, PRQ reduces quantization errors through iterative stacked layers but uniquely decomposes whole-body motion features into anatomically meaningful joint groupings, quantizing them as discrete part-level codes. 
This design enables \texttt{\textbf{Being-M0.5}} to achieve real-time generation through efficient frame-by-frame motion code decoding, significantly outperforming previous approaches in computational efficiency. 

Our key contributions are as follows:
\begin{itemize}
    \item \textbf{Large-scale multimodal dataset}: We present \textbf{\texttt{HuMo100M}}, the largest multimodal motion dataset to date, comprising 5 million motion sequences and 100 million multi-task instructional instances with fine-grained, long-form, and part-level annotations.
    \item \textbf{Real-time controllable model}: We propose \texttt{\textbf{Being-M0.5}}, a highly controllable VLMM that achieves superior performance compared to existing models, accompanied by comprehensive analysis of key design choices and architectural innovations.
    \item \textbf{Part-aware motion control}: We develop part-aware residual quantization that leverages \texttt{\textbf{HuMo100M}}'s part labels to enable fine-grained motion control, addressing a critical limitation in current VLMMs.
\end{itemize}

%% file: sec/2_relatedwork.tex
\section{Related Work}
\label{sec:related_work}

\noindent\textbf{Human Motion Generation.}
Human motion generation is typically categorized by the control signals employed, including text descriptions~\cite{guo2022generating,petrovich2022temos}, action labels~\cite{ahuja2019language2pose}, keyframe poses~\cite{zhang2024motiongpt}, and incomplete motion sequences~\cite{wang2024motiongpt}. 
Early deterministic text-to-motion (T2M) methods often produced blurry or unrealistic results~\cite{fragkiadaki2015recurrent,gopalakrishnan2019neural}, while subsequent approaches leveraged stochastic techniques such as variational autoencoders (VAEs)~\cite{aliakbarian2020stochastic} and generative adversarial networks (GANs)~\cite{wang2020learning} to address these limitations.
Recent advances~\cite{jiang2023motiongpt,wang2024motiongpt} have integrated large language models (LLMs) to better interpret human intent and generate more contextually appropriate motions. 
Notable examples include MotionGPT~\cite{zhang2024motiongpt} and its variants, which demonstrate improved motion understanding through language model integration.
MotionChainn~\cite{jiang2024motionchain} extends this paradigm by enabling multi-turn conversational motion generation, while MotionLLM~\cite{chen2024motionllm} provides a unified framework for motion understanding, captioning, and reasoning. 
Additional efforts such as LMM~\cite{zhang2024large} and recent multimodal approaches~\cite{luo2024m} have explored human-centric video understanding to enhance motion generation capabilities.
Despite these advances, existing research often overlooks critical aspects of motion generation controllability and fails to achieve an optimal balance between model performance and computational efficiency, particularly for real-time applications.

\noindent\textbf{Vision-Language Models.}
Our work aims to develop a real-time, controllable vision-language-motion model (VLMM) built upon established vision-language model (VLM) architectures~\cite{alayrac2022flamingo,li2023blip}. 
Contemporary VLMs~\cite{alayrac2022flamingo,li2023blip} leverage both visual and textual contexts to generate coherent responses, typically consisting of a visual encoder, feature projection layers, and an LLM backbone. 
A fundamental challenge in VLM development lies in effectively bridging the semantic gap between visual and textual modalities.
LLaVA~\cite{liu2024visual} exemplifies successful cross-modal integration by employing linear projections to map visual features into the LLM's representational space, combined with visual instruction tuning to create versatile visual assistants.
This paradigm has inspired numerous extensions, including Video-LLaVA~\cite{lin2023video} for temporal visual understanding and specialized models for diverse vision-language tasks beyond traditional visual question answering~\cite{chen2023shikra,lai2024lisa}.
In this work, we extend the input modality space to encompass human motion data while adopting proven VLM architectures as our foundational framework.

\noindent\textbf{Motion Tokenization.}
Effective motion representation is crucial for high-quality generation, with vector quantization (VQ)~\cite{van2017neural} serving as a foundational approach for robust human motion encoding. 
Recent advances have introduced several sophisticated quantization techniques that significantly improve motion representation quality. These include residual quantization (RQ)~\cite{guo2024momask} for iterative refinement, hierarchical quantization (HQ)~\cite{you2022locally} for multi-scale representation, lookup-free quantization (LFQ)~\cite{yu2023language} for efficiency, and finite scalar quantization (FSQ)~\cite{mentzer2023finite} for enhanced expressiveness.
Emerging research has begun exploring part-level motion tokenization to enable more granular control.
Some approaches~\cite{chen2024language} partition the body into upper and lower segments, while others~\cite{lu2023humantomato,zhang2024motiongpt} focus on specific components such as body and hand motions. 
However, these methods typically lack comprehensive independent limb control capabilities and corresponding textual annotations or evaluation benchmarks.
This limitation motivates our development of a novel part-aware tokenization approach that enables real-time, fine-grained control over individual body parts while maintaining computational efficiency.

%% file: sec/3_method.tex
\section{Being-M0.5}

\begin{figure*}[t]
\centering
    \includegraphics[width=1\linewidth]{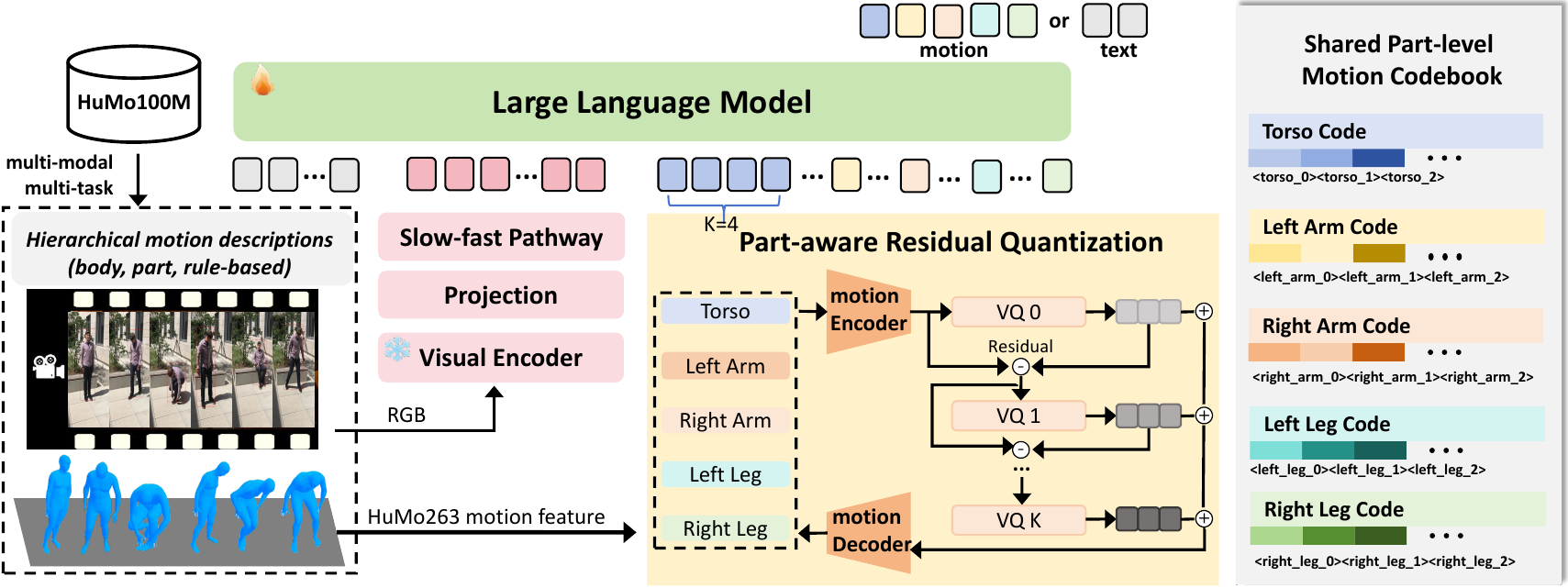}
\caption{\textbf{Model Illustration.} 
\texttt{\textbf{Being-M0.5}} supports multi-modal inputs/outputs, built on a 7B LLM backbone. It employs SigLIP+2MLP for visual encoding and projection with a slow-fast strategy, alongside part-aware residual quantization for motion tokenization.}
\label{fig:model_structure}
\end{figure*}

We present \texttt{\textbf{Being-M0.5}}, a 7B-parameter vision-language-motion model (VLMM) trained on 5 million human motion sequences and 100 million motion instructional instances, as illustrated in Figure~\ref{fig:model_structure}.
Despite recent progress in motion generation, critical questions regarding real-time, controllable VLMMs remain largely unresolved. 
To address these challenges, we provide a comprehensive model overview (Section \ref{sec:overview}), followed by detailed analyses of controllable motion generation strategies
(Section \ref{sec:method_control}) and real-time design considerations (Section \ref{sec:method_real_time}).

\subsection{Overview of VLMM}
\label{sec:overview}
Our VLMM is built on the LLaVA-video-7B framework~\cite{zhang2024video}.
Following established VLM designs~\cite{chen2024internvl,liu2023visual}, our model comprises three core components: a 400M visual encoder (SigLIP~\cite{zhai2023sigmoid}), a 2-layer MLP for visual feature projection, and a 7B LLaMA-2-chat backbone~\cite{touvron2023llama}. 
To efficiently process extended video sequences, we adopt a slow-fast strategy that reduces visual token complexity while preserving temporal information.
Our model conceptualizes human motion as a structured language with its own vocabulary and syntax.
Given a motion sequence $m_{1:T}$, where $m_i \in \mathbbm{R}^D$ represents a $D$-dimensional motion feature vector and $T$ denotes the sequence length, we employ a motion tokenizer $\mathcal{Q}$ to quantize the continuous motion data into discrete tokens. 
We extend the base VLM vocabulary with $K$ additional motion-specific codes and introduce special boundary tokens \texttt{<MOT>} and \texttt{</MOT>} to delineate motion sequence spans within the text.
Each training instance follows an instruction-following format $\{\mathcal{X}_Q, \mathcal{X}_A\}$, representing a User-VLMM dialogue pair. 
We curate paired and interleaved vision, language, and motion data from web videos to support diverse tasks including text-to-motion generation, motion prediction, and multimodal motion understanding.
Given a query $\mathcal{X}_Q$, the VLMM generates response $\mathcal{X}_A=\{y_1, y_2,...,y_n\}$, where each $y_i$ can be a text token or motion code.
The model is optimized using the standard next-token prediction objective via negative log-likelihood:

\begin{equation}
\mathcal{L}(\Theta)=-\sum_{j=1}^{L} \log P_{\Theta}(y_j \mid \mathcal{X}_Q, \hat{y}_{1:j-1}).
\end{equation}

Our VLMM training follows a three-stage curriculum: (1) motion-text alignment to establish correspondence between motion tokens and the LLM's representational space; (2) vision-text-motion alignment to integrate all three modalities into a unified framework; and (3) motion instruction tuning to enhance responsiveness to diverse user commands and improve controllability.

\subsection{Controllable Motion Generation}
\label{sec:method_control}

Prior research has largely overlooked the controllable potential of VLMMs, significantly limiting their practical applicability. 
This work addresses this critical gap through two complementary approaches: comprehensive multi-task pretraining and novel architectural innovations.

\subsubsection{Multi-Task Motion Pretraining}

We define ``controllability'' through five fundamental dimensions and develop these capabilities through systematic data curation and the design of specialized instructional tasks.

\noindent\textbf{Random Instruction Control.} 
Most existing VLMMs exhibit limited ability to handle arbitrary user commands effectively, constraining their real-world utility.
While some studies~\cite{zhang2024large} have explored motion instruction tuning, we significantly enhance command responsiveness by creating a comprehensive instruction template library (e.g., ``Show me how to perform \texttt{<CAPTION>}.'', where \texttt{<CAPTION>} represents the motion description) and introducing the Instruct-to-Motion (I2M) task for robust natural language understanding.

\noindent\textbf{Random Pose Initialization Control.}
A practical VLMM should generate coherent motion from arbitrary initial poses to emulate human adaptability, rather than being constrained to fixed configurations like the canonical T-pose.
However, current VLMMs struggle with this capability due to limited training data diversity. 
To address this limitation, we develop the Motion Prediction and In-between (MPI) task, where we randomly extract prior, intermediate, or posterior segments from motion sequences and challenge the VLMM to predict the remaining portions. 
This approach necessitates million-scale datasets, motivating our development of \texttt{\textbf{HuMo100M}} to enable learning from diverse initial pose configurations.

\noindent\textbf{Long-Term Motion Control.} 
Human activities typically unfold as seamless, continuous sequences, and practical VLMMs should similarly generate extended motion sequences rather than isolated snippets.
To achieve this capability, we leverage the concatenated long-form motion sequences within \texttt{\textbf{HuMo100M}} and introduce the Instruct-to-LongMotion (I2LM) task, enabling the generation of temporally coherent, extended motion sequences.

\noindent\textbf{Unseen Motion Control.}
Existing datasets lack sufficient scale to ensure robust generalization to novel motion patterns not encountered during training. 
To address this limitation, we expand motion data coverage through large-scale web video collection and multi-task training design. 
Leveraging \texttt{\textbf{HuMo100M}}'s unprecedented scale with millions of motion instances, our VLMM demonstrates strong generalization capabilities for previously unseen actions. 
We introduce the Instruct-to-Unseen (I2U) task to systematically evaluate this generalization capability.

\noindent\textbf{Random Part Control.}
Advanced VLMMs should enable precise control over specific anatomical regions (e.g., ``kicking with the left leg'' or ``waving the right hand'').
Previous approaches often fail due to insufficient part-level supervision, even when employing part-aware motion encoders~\cite{chen2024language}.
Utilizing our comprehensive part-level annotations, we propose the Instruct-to-PartMotion (I2PM) task, which challenges the model to generate motion for specific body regions while maintaining overall motion coherence and realism.

\subsubsection{Part-aware Residual Quantization (PRQ).}
Given a motion sequence $m_{1:T} \in \mathbb{R}^{T\times D}$, our PRQ first decomposes each motion feature $m_i$ into part-specific features $m_{i,j} \in \mathbb{R}^{d}$, where $j \in \{1, 2, \ldots, p\}$, $d$ represents the part feature dimension, and $p = 5$ corresponds to anatomically meaningful body regions: left arm, right arm, left leg, right leg, and torso. Note that elements across different $m_{i,j}$ may overlap due to shared joint representations.
PRQ then encodes these part features into a latent vector sequence $\tilde{b}_{1:n;1:p}$ using a shared encoder with downsampling ratio $n/T$. Each latent vector $\tilde{b}_{i,j}$, where $i \in \{1, 2, \ldots, n\}$, is quantized by identifying its nearest neighbor in a shared motion codebook $\mathbb{C}$, producing the discrete code sequence $b_{1:n;1:p}$.
Following residual quantization principles, PRQ represents a latent sequence $\tilde{b}_{1:n;1:p}$ as $K+1$ ordered code sequences across $K+1$ quantization layers:

\begin{equation}
\text{PRQ}(\tilde{b}_{1:n;1:p}) = [b_{1:n;1:p}^k]_{k=0}^K
\end{equation}

where $b^k$ denotes the code sequence at quantization layer $k$.
To reconstruct motion $\tilde{m}$, PRQ's decoder maps the quantized latent sequence $\tilde{b}_{1:n;1:p}$ back to part-level motion space $\tilde{m}_{1:n;1:p}$, then aggregates each $\tilde{m}_{i,1:p}$ to restore the complete motion feature $\tilde{m}_i$ by selecting corresponding elements from each part feature.
During tokenization, starting from the initial residual $r^0 = \tilde{b}$, PRQ iteratively computes $b^k$ as the optimal approximation of residual $r^k$, updating the residual $r^{k+1}$ as:

\begin{equation}
b^k=\mathcal{Q}(r^k), \quad r^{k+1}=r^k-b^k
\end{equation}

This residual processing operates independently for each body part. The final latent sequence approximation for part $j$ is: $b_j = \sum_{k=0}^K b_j^k$.
Similar to standard RQ, PRQ is trained using motion reconstruction and latent embedding objectives across all quantization layers:

\begin{equation}
\mathcal{L} = \sum_{j=1}^p \|m_j - \tilde{m}_j\|_1 + \|m - \tilde{m}\|_1 + \beta \sum_{k=1}^K \sum_{j=1}^p \|r^k_j - \text{sg}[b^k_j]\|_2^2
\end{equation}

where $\text{sg}[\cdot]$ denotes the stop-gradient operation and $\beta$ weights the embedding loss.
Rather than using monolithic codes to represent entire body configurations, PRQ employs part-specific codes that enable independent control over individual anatomical regions. 
Additionally, PRQ effectively expands codebook capacity without increasing storage requirements.
If part $j$ references $u_j$ distinct codes, the total number of representable motion configurations becomes $\prod_{j=1}^p u_j$.
Compared to prior work~\cite{chen2024language} that focuses solely on upper and lower body partitioning, PRQ introduces three key innovations: (1) finer-grained control over five distinct anatomical regions, 
(2) rich part-level textual supervision enabling precise instruction following, and 
(3) shared motion codebook that enables joint representations across limbs, reducing inter-joint coordination errors.
Detailed specifications (e.g., part feature definitions and anatomical mappings) are provided in Appendix~\ref{app:prq}.

\subsection{Design Choices for Real-time Generation}
\label{sec:method_real_time}

Achieving real-time motion generation requires careful consideration of computational efficiency across all model components. 
We systematically evaluate key design choices that impact inference speed while maintaining generation quality.
 
\noindent\textbf{LLM Backbone Selection.}
To optimize the performance-efficiency trade-off, we conduct comprehensive experiments across multiple LLM architectures. 
Our analysis reveals that smaller models (e.g., GPT-2) lack sufficient capacity to effectively interpret complex human intent, while models exceeding 13B parameters suffer from prohibitive inference latency that precludes real-time deployment. 
Based on extensive evaluation, we select the 7B-parameter LLaMA-2 as \texttt{\textbf{Being-M0.5}}'s backbone, which provides optimal balance between language understanding capability and computational efficiency.

\noindent\textbf{Motion Feature.}
The HM3D263 format~\cite{guo2022generating} represents a widely adopted feature representation in contemporary motion generation research, incorporating relative joint positions, velocities, 6D rotations of key joints, and foot contact information.
However, a critical limitation of HM3D263 lies in its indirect computation of rotation information from joint position data via inverse kinematics (IK). 
This approach not only introduces information loss through the intermediate transformation but also imposes significant computational overhead and latency due to the iterative IK solving process, making it unsuitable for real-time applications.
To address these limitations, we propose HuMo263, a novel motion feature based directly on SMPL model parameters~\cite{loper2023smpl}. 
HuMo263 comprises: (1) relative 6D rotations of key joints (126 dimensions), (2) root node parameters (4 dimensions: angular velocity, xz-velocity components, and y-height), (3) redundant joint position information derived from SMPL forward kinematics (63 dimensions), and (4) foot contact states (4 dimensions). Unlike HM3D263, HuMo263 directly preserves rotation information from the SMPL model output, eliminating information degradation and computational delays associated with IK calculations.

\noindent\textbf{Visual Resolution and Temporal Duration.}
Visual input resolution significantly impacts computational efficiency. 
Through systematic comparison across multiple resolutions, we observe that higher resolutions provide marginal quality improvements while substantially increasing computational burden. For optimal real-time performance, \texttt{\textbf{Being-M0.5}} operates at 224$\times$224 resolution and processes up to 64 input frames, striking an effective balance between visual fidelity and inference speed.

\noindent\textbf{Tokenization Architecture and Decoding Strategy.}
While stacked quantization layers (e.g., residual quantization) enhance motion representation accuracy, they linearly increase computational complexity. 
Our empirical analysis demonstrates that 4 PRQ layers achieve optimal quality-efficiency balance. 
Critically, unlike standard RQ approaches that employ layer-by-layer generation --- delaying output completion until the final token --- PRQ utilizes frame-by-frame decoding strategy. 
This architectural choice enables progressive motion output generation, significantly improving system responsiveness for real-time applications.

%% file: sec/4_dataset.tex
\begin{figure*}[ht]
\vspace{.3cm}
\centering
\includegraphics[width=1\linewidth]{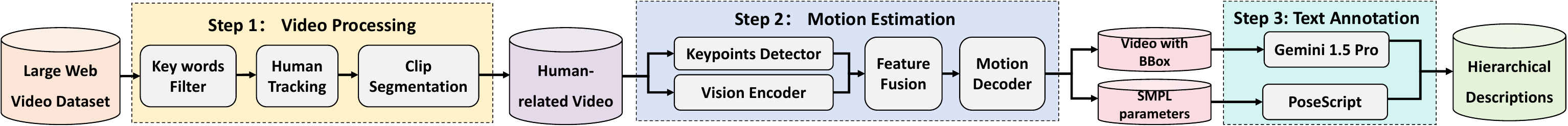}
\caption{\textbf{Illustration of data pipeline.} We introduce \texttt{\textbf{HuMo100M}}, the largest multimodal dataset to date, featuring over 5 million motion sequences, paired visual clips, threefold more hierarchical and part-level textual descriptions, and 100 million multi-task instruction instances. 
}
\label{fig:dataset_pipline}
\end{figure*}

\section{The HuMo100M Dataset}
While scaling training data has proven fundamental to the success of large multimodal models, this strategy faces significant challenges in motion generation due to the chronic scarcity of high-quality motion datasets~\cite{lu2024scamo}. 
To address this critical limitation, we introduce \texttt{\textbf{HuMo100M}}, the largest multimodal human motion dataset to date, comprising over 5 million motion sequences and 100 million instructional instances. Figure~\ref{fig:dataset_pipline} illustrates our construction pipeline overview. 
Additional details are provided in Appendix~\ref{app:dataset}.

\subsection{Data Curation}
To construct HuMo100M, we collect over 20 million videos from publicly available datasets and online platforms, employing a systematic multi-stage filtering and processing pipeline to ensure high-quality human motion extraction. 
Our pipeline comprises sequential stepts designed to maximize data quality and relevance.

To begin with, we perform initial content filtering by analyzing video metadata, discarding content whose textual descriptions lack human-related terminology (e.g., people, human, person, man, woman). 
This preliminary step efficiently eliminates irrelevant content at scale.
While keyword-based filtering ensures human-related content at the video level, many temporal segments within these videos lack actual human presence. 
To address this challenge and enhance dataset quality, we implement YOLO-based tracking~\cite{redmon2016you} to monitor every individual throughout each video sequence, enabling precise identification and extraction of human-centric segments.
We subsequently perform video segmentation based on tracking results to preserve trajectory integrity and maintain spatiotemporally consistent motion sequences. 
This ensures that extracted motion data represents continuous, coherent human activities rather than fragmented or interrupted sequences.

Occlusion and motion blur represent common challenges in real-world human motion videos that can significantly degrade motion estimation quality. To mitigate these issues, we develop a comprehensive quality assessment with two complementary components.
First, we apply a pre-trained 2D keypoint detector to extract skeletal landmarks for each human subject, filtering detection results based on confidence thresholds. 
We utilize the number of successfully detected high-confidence keypoints as the primary metric for assessing potential occlusions.
Motion sequences are flagged as significantly occluded when the number of visible keypoints falls below a predetermined minimum threshold, ensuring only high-visibility sequences are retained.
Second, short motion sequences contain insufficient temporal context and often exhibit reduced estimation accuracy due to limited observational data. We implement an effective sequence length filtering mechanism that removes motion clips containing insufficient frames, ensuring that retained sequences provide adequate temporal information for reliable motion analysis and generation.

After the above filtering steps, we extract SMPL parameters using WHAM~\cite{shin2024wham} to regress accurate 3D human motion in world coordinates, providing the foundational motion representation for our dataset.
Finally, we enhance motion quality using the RL-based policy PHC~\cite{luo2023perpetual}, following the approach established by~\cite{wang2024quo}.

\subsection{Key Advantages}

Compared to existing datasets~\cite{wang2024quo,lu2024scamo,xu2024motionbank}, \textbf{\texttt{HuMo100M}} offers three distinctive advantages that significantly advance the state of motion generation research and enable more sophisticated controllable generation capabilities.

\subsubsection{Part-Level Text Description} 
Existing motion datasets predominantly rely on body-level descriptions that provide high-level summaries of overall movement patterns in concise sentences. 
However, these coarse-grained annotations lack fine-grained details about individual anatomical regions, severely limiting their utility for precise part-level motion control --- a critical capability for practical applications.
To address this fundamental limitation, we augment each motion sequence with comprehensive limb-specific descriptions alongside traditional whole-body annotations~\cite{guo2022generating,lin2024motion}.
Our part-level descriptions are anatomically structured: upper body annotations detail arm, shoulder, and torso movements, while lower body descriptions focus on leg, hip, and foot actions. 
These annotations are generated using Gemini-1.5-Pro~\cite{team2024gemini} with carefully engineered prompts and further enhanced through PoseScript integration~\cite{delmas2024posescript} for improved semantic accuracy.
Additionally, we incorporate rule-based descriptions derived from \textit{posecodes} that extract semantic spatial relationships between joints, such as ``the left hand is positioned below the right hand'' or ``the right foot is extended forward of the left knee''.
This approach captures precise inter-joint spatial interactions that are crucial for fine-grained motion control.
This multi-level annotation strategy enables sophisticated part-level motion control and allows VLMMs to prioritize high-confidence keypoint detections while intelligently disregarding occluded or unreliable regions, substantially improving motion data quality extracted from challenging web video sources.

\subsubsection{Long-term Motion Sequence}
\label{sec:long-motion}
Current motion datasets are predominantly composed of short-duration motion clips, typically lasting only a few seconds. 
While large language models demonstrate remarkable capability in generating extended textual sequences, this temporal limitation in motion datasets severely constrains models' ability to generate realistic long-sequence motions from textual descriptions, which is a critical requirement for practical applications involving complex, multi-stage activities.
To bridge this temporal gap, we develop two motion concatenation methods that integrate shorter motion sequences into longer, spatiotemporally consistent sequences.

\noindent\textbf{Interpolation-based Concatenation.}
We implement a sophisticated three-stage alignment process to ensure seamless motion concatenation. 
First, we perform orientation alignment between motion sequences to establish consistent directional coherence. 
Second, we conduct global coordinate alignment through precise spatial translation.
Lastly, we establish smooth transitions by selecting a neutral standing pose as a reference state and applying spherical linear interpolation (Slerp) between motion sequences and this reference pose.
This method ensures natural transitions where characters return to standardized postures between activities, maintaining both continuity and physical plausibility throughout extended sequences.

\noindent\textbf{Learning-based Concatenation.} 
We introduce a specialized in-between motion prediction model designed to generate contextually appropriate transitions between disparate motion sequences.
Our approach leverages a pre-trained text-to-motion model, fine-tuned specifically for transition generation using a strategic masking approach that occludes approximately 50\% of motion sequences during training, enabling robust motion completion capabilities.

During inference, our hybrid framework operates through a two-stage process: the interpolation-based method first generates initial keyframes to bridge sequences, which are then refined by the learning-based model to produce smooth, contextually consistent transitions.
This approach combines the reliability of rule-based methods with the adaptability of data-driven approaches, ensuring generated transitions are both physically plausible and semantically coherent.

\subsubsection{Text-Aligned Visual Clips}
While several unified motion models~\cite{li2024unipose,chen2024language,zhang2024large} have begun incorporating visual information, the full potential of vision-language alignment in motion generation remains largely unexplored.
We argue that visual cues provide limited additional value for high-quality motion capture datasets (e.g., HumanML3D) but become increasingly beneficial when training on lower-quality motions extracted from web videos.
This phenomenon occurs because even when motion estimation from web videos produces noisy or incomplete data, VLMMs can leverage weak supervision signals from text-aligned visual clips, utilizing rich contextual visual information to enhance motion understanding and compensate for motion estimation errors. 
The visual modality provides crucial contextual cues about environment, object interactions, and movement dynamics that may be lost or corrupted during automated motion extraction processes.
Furthermore, as a unified multimodal framework, integrating visual clips enables VLMMs to perform motion estimation tasks directly from video input, effectively mimicking observed human actions and significantly broadening the scope of real-world applications beyond text-conditioned generation alone.

%% file: sec/5_experiments.tex
\section{Experiments}

\subsection{Experimental Setup}

\subsubsection{Datasets}

To comprehensively evaluate \texttt{\textbf{Being-M0.5}}'s performance across diverse motion generation tasks, we conduct experiments on both established benchmarks and newly constructed datasets that leverage our HuMo100M dataset. 
Our evaluation framework encompasses nine distinct tasks, each designed to assess specific aspects of controllable motion generation.

\begin{itemize}[leftmargin=1.5em]

\item \textbf{Text-to-Motion (T2M):} 
We first evaluate on two standard benchmarks: HumanML3D~\cite{guo2022generating} (14,616 motions with 44,970 text descriptions) and KIT-ML~\cite{plappert2016kit} (3,911 motions with 6,278 descriptions). 
Both datasets follow an train (80\%) / validation (5\%) / test (15\%) split.
In addition, we create a large-scale testbed HuMo-T2M by sampling 200,000 high-quality motion-text pairs from HuMo100M, enabling evaluation on more diverse scenarios than traditional benchmarks.

\item \textbf{Instruct-to-Motion (I2M):} 
We construct a comprehensive benchmark HuMo-I2M containing one million high-quality motion-instruction pairs sampled from HuMo100M, maintaining the standard 80\%/5\%/15\% split ratio. 
This benchmark evaluates the model's ability to follow diverse natural language instructions.

\item \textbf{Instruct-to-Unseen (I2U):}
To assess generalization capabilities, we curate 200,000 novel motion sequences to build HuMo-Unseen from HuMo100M that are completely absent from training data, providing rigorous evaluation of the model's ability to generate previously unseen motion patterns.

\item \textbf{Instruct-to-PartMotion (I2PM):} 
We develop a specialized benchmark named HuMo-I2PM for part-level control evaluation, containing 200,000 instances with detailed anatomical region annotations (150,000 test, 50,000 validation). This benchmark specifically assesses fine-grained body part control capabilities.

\item \textbf{Instruct-to-LongMotion (I2LM):} 
For extended sequence generation evaluation, we create HuMo-I2LM with 500,000 concatenated motion sequences using our proposed concatenation methods, following standard split ratios. 
This benchmark evaluates temporal consistency and long-term motion coherence.

\item \textbf{Motion Prediction and In-between (MPI):}
We supplement standard AMASS~\cite{mahmood2019amass} and 3DPW~\cite{von2018recovering} evaluations with HuMo-MPI, a 200,000-sample dataset extracted from HuMo100M. 
Following established protocols~\cite{zhang2024large}, we focus on six anatomical regions (spine, arms, legs, head) while excluding global translation and facial expressions.

\item \textbf{Action-to-Motion:} 
We utilize the UESTC dataset~\cite{ji2018large} for action-conditioned motion generation, following established evaluation protocols~\cite{zhang2024large}.

\item \textbf{Motion Reconstruction:} 
We evaluate reconstruction capabilities across three datasets: HumanML3D, MotionX~\cite{lin2023motion}, and HuMo100M, providing assessment of motion encoding and decoding quality.
    
\item \textbf{Motion-to-Text:} 
We assess motion captioning capabilities using the HumanML3D benchmark, evaluating the model's ability to generate descriptive text from motion sequences.

\end{itemize}

\subsubsection{Evaluation Metrics} 
We employ task-specific evaluation metrics to comprehensively assess \texttt{\textbf{Being-M0.5}}'s performance across different motion generation and understanding capabilities.
For motion generation, we utilize established motion evaluation metrics that assess both motion quality and text-motion alignment: 
(1) Fréchet Inception Distance (FID) measures the distributional divergence between generated and ground truth motion sequences by comparing high-level feature distributions.
(2) R-Precision (top-1, top-2, top-3) evaluates text-motion semantic alignment by measuring retrieval accuracy when ranking ground-truth text descriptions among candidates, assessing how well generated motions correspond to their textual prompts. 
(3) Multimodal Distance (MM-Dist) quantifies the distance between text and motion features within a shared embedding space, measuring cross-modal alignment quality.
For reconstruction and prediction tasks, we employ kinematic accuracy measures: 
Mean Per Joint Position Error (MPJPE) calculates the average joint position error in millimeters between predicted and ground-truth poses, providing precise quantitative assessment of motion fidelity. 
We also utilize FID for these tasks to evaluate the overall quality of reconstructed motion distributions.

\subsubsection{Implementation Details}
Our part-aware residual VQ-VAE (PRQ) employs residual blocks for both motion encoder and decoder components, with a temporal downsampling rate of 4$\times$. 
The PRQ codebook contains 1,024 entries with 512-dimensional embeddings across 4 quantization layers. 
These discrete motion codes are seamlessly integrated into the LLM vocabulary as additional tokens.
For optimal performance-efficiency balance, we utilize LLaMA-2-7B~\citep{touvron2023LLaMA2} as our language model backbone. 
As a baseline comparison, we implement a standard VQ motion tokenizer with identical codebook specifications.
The motion tokenizer training employs a batch size of 256 with a learning rate of 1e-4 across 300,000 iterations. 
Our VLMM training follows a three-stage curriculum:
(1) motion-text alignment with full parameter fine-tuning on 16$\times$A800 GPUs with batch size 2,048 for 50 epochs, using learning rate 2e-5.
(2) vision-text-motion alignment with batch size 128 for 5 epochs. 
(3) motion instruction tuning with batch size 128 for one epoch to enhance instruction-following capabilities.

\input{tables/01_sota_t2m}

\subsection{Multi-Task Benchmarking}
We conduct comprehensive comparisons between \texttt{\textbf{Being-M0.5}} and state-of-the-art methods across multiple benchmarks to evaluate performance across diverse motion generation tasks. 
For fair evaluation, all baseline methods utilize identical motion data within each benchmark, except \texttt{\textbf{Being-M0.5}}$^*$, which leverages the full HuMo100M dataset to demonstrate the benefits of large-scale training.

\noindent\textbf{Text-to-Motion (T2M).}
Text-to-motion represents a fundamental benchmark for evaluating motion understanding and generation capabilities. Table~\ref{tab:sota_t2m} presents comprehensive comparison with all models trained on the HumanML3D dataset.
To isolate the contribution of our proposed components, we first evaluate \texttt{\textbf{Being-M0.5}} using standard vector quantization (VQ). 
Even with this baseline configuration, our model demonstrates superior performance compared to existing methods~\cite{zhang2024motiongpt,zhang2024large} in both generation accuracy and motion fidelity.
The integration of our proposed part-aware residual quantization (PRQ) yields substantial improvements, reducing the FID score from 0.141 to 0.056 --- a 60\% improvement that indicates significantly enhanced motion quality and realism.

\begin{wrapfigure}{r}{.5\textwidth}
\vspace{-.1cm}
\centering    
\includegraphics[width=0.45\textwidth]{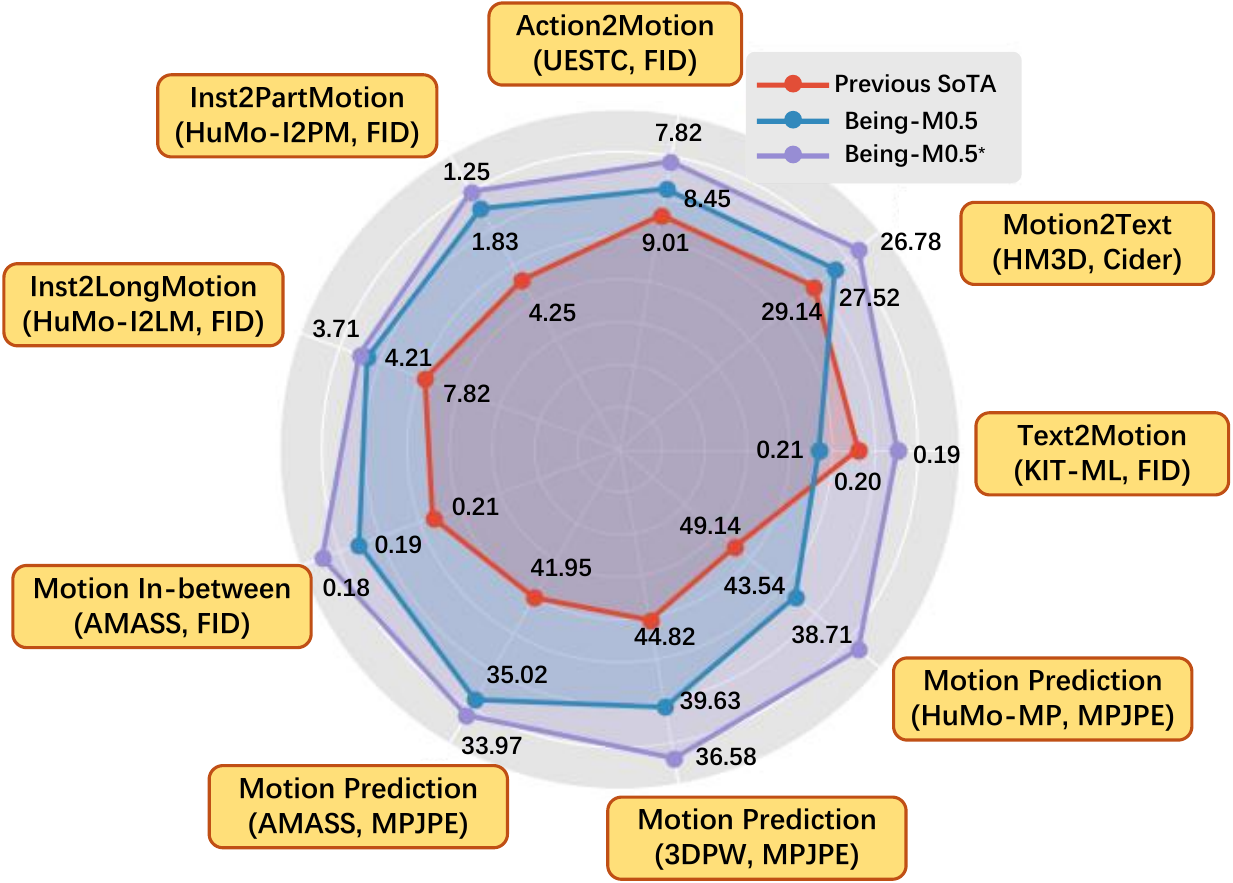}
\caption{
Comparison with previous SoTAs across nine benchmarks. Here, \texttt{\textbf{Being-M0.5}}$^{*}$ indicates the model trained on the complete HuMo100M dataset. 
For new benchmarks introduced in this paper (e.g., I2PM), we adopt MotionGPT~\cite{wang2024motiongpt} as the baseline.}
\label{fig:sota_ridar} 
\vspace{-.2cm}
\end{wrapfigure}

We further benchmark \texttt{\textbf{Being-M0.5}} against recent models employing sophisticated quantization variants: SCaMo~\cite{lu2024scamo} with Finite Scalar Quantization (FSQ)~\cite{lu2024scamo} and MoMask~\cite{guo2024momask} with Residual Quantization (RQ)~\cite{lee2022autoregressive}. 
Despite these methods utilizing substantially larger codebooks (64K entries) or deeper quantization layers, our model consistently outperforms these methods, demonstrating the efficiency and effectiveness of our PRQ approach being a more compact representations.
Additional results on the KIT-ML benchmark are presented in Figure~\ref{fig:sota_ridar}, the results show consistent performance improvements across different evaluation scenarios and confirming the robustness of our approach.

\noindent\textbf{Instruct-to-Motion (I2M).}
Unlike T2M, the I2M task requires generating motion sequences from arbitrary human commands, thereby testing the model's capability to respond effectively to real-world instructions. 
As demonstrated in Table~\ref{tab:sota_i2m}, MoMask exhibits significantly degraded performance on I2M compared to T2M, highlighting the critical need for LLM integration to accurately interpret human intent.
When compared to other LLM-based approaches such as MotionGPT~\cite{wang2024motiongpt}, our model achieves better results.
We attribute this improvement to two key factors: our enhanced motion instruction tuning methodology and the introduction of PRQ.

\input{tables/02_sota_i2m}

\noindent\textbf{Instruct-to-Unseen (I2U).}
To establish this benchmark, we curate 200K novel motion sequences, termed HuMo-Unseen, which are excluded from all training datasets. 
As illustrated in Table~\ref{tab:unseen}, performance on HuMo-Unseen demonstrates consistent improvement with data scaling from HumanML3D to HuMo100M, emphasizing the critical importance of large-scale data for achieving robust generalization to unseen motion patterns.

\input{tables/03_unseen}

\noindent\textbf{Instruct-to-PartMotion (I2PM).}
Part-level motion control represents a fundamental aspect of controllability that has received limited attention in prior research. 
While some studies have investigated part-aware motion quantization, they suffer from insufficient fine-grained annotations and lack comprehensive benchmarks for validation.
To address this limitation, we leverage Gemini-1.5 Pro to collect motion sequences with specific part-level instructions (e.g., ``raise your left arm'') and construct the HuMo-I2PM benchmark. 
As presented in Figure~\ref{fig:sota_ridar},  \texttt{\textbf{Being-M0.5}} substantially outperforms MotionGPT, primarily attributed to our part-aware motion encoding strategy. 
Detailed analysis is provided in Section~\ref{sec:motion_quant}.

\noindent\textbf{Instruct-to-LongMotion (I2LM).}
This task evaluates the model's capacity for generating extended, continuous motion sequences. 
We construct this benchmark by concatenating individual motion components with complex instructions (e.g., ``Salute with your left hand and perform ballet''). 
Consistently, \texttt{\textbf{Being-M0.5}} shows significant improvement over baseline methods.

Following Zhang~\etal~\cite{zhang2024large}, we conduct additional evaluations across multiple benchmarks, including motion prediction on AMASS~\cite{mahmood2019amass}, 3DPW~\cite{von2018recovering} and HuMo-MPI; motion in-between on AMASS; action-to-motion on UESTC~\cite{ji2018large} and motion-to-text on HumanML3D (HM3D)~\cite{guo2022generating}.
The comprehensive comparison results are presented in Figure~\ref{fig:sota_ridar}.

\input{tables/04_quantization}

\subsection{Main Analysis}

\subsubsection{Motion Quantization Configuration}
\label{sec:motion_quant}

In Table~\ref{tab:motion_quant}, we present a comprehensive comparison between our part-aware residual quantization (PRQ) and existing motion tokenization methods.
First, PRQ substantially outperforms lookup-free approaches such as 2D-LFQ and FSQ while utilizing a codebook that is only 1.5\% of their size. 
This performance advantage becomes even more pronounced on the large-scale \textbf{\texttt{HuMo100M}} testbed, demonstrating PRQ's superior generalization capabilities.
We attribute this effectiveness to our part decomposition strategy, which enhances codebook capacity without proportional increases in memory requirements.
Furthermore, PRQ surpasses RQ-VAE performance even when employing fewer quantization layers, owing to its part-level encoding mechanism. 
As previously introduced, our design incorporates shared joints across different body parts to enhance inter-part connectivity and reduce joint reconstruction error. 
To validate the effectiveness of this shared joint strategy, we conduct an ablation study comparing PRQ against its variant without shared joints (PRQ w/o Shared).
The results consistently demonstrate that PRQ outperforms this ablated variant, confirming the importance of our shared joint design.

\subsubsection{Time Efficiency}
\begin{figure}[ht]
\centering    
\includegraphics[width=0.6\textwidth]{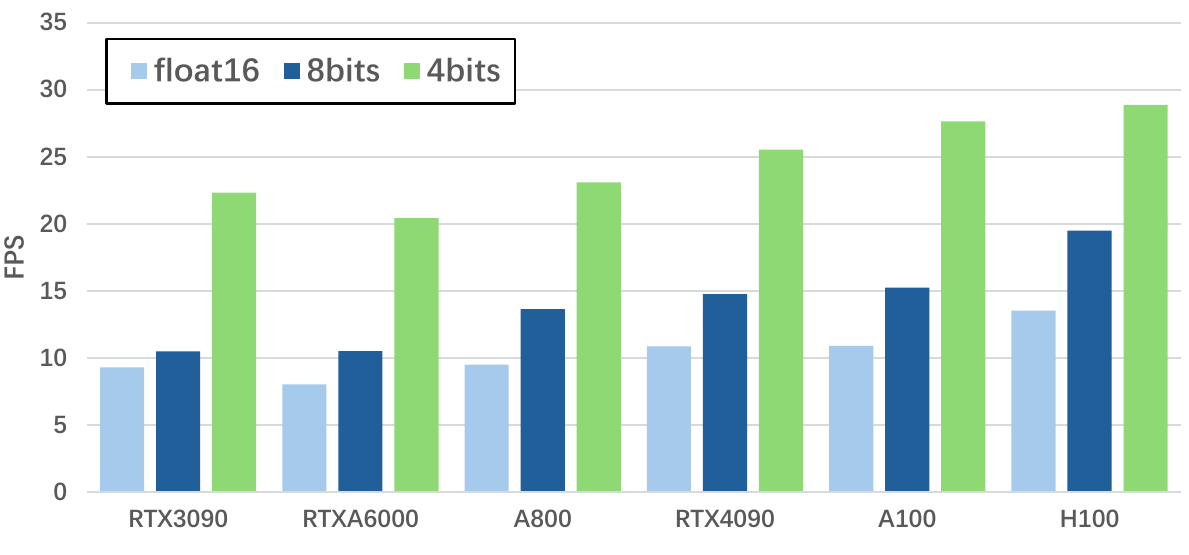}
\caption{
\texttt{\textbf{Being-M0.5}} inference speed across different GPUs. We accelerate motion generation via modern LLM inference framework llama.cpp~\cite{llamacpp}. Our model achieves real-time inference speed using the 7B-parameter LLaMA backbone.
}
\label{fig:time_efficient} 
\end{figure}

The 7B-parameter \texttt{\textbf{Being-M0.5}} model exhibits higher memory requirements during inference compared to smaller models (<1B parameters)~\cite{zhang2024large}.
Figure~\ref{fig:time_efficient} illustrates its inference performance across different GPU architectures.
With 4-bit quantization, \texttt{\textbf{Being-M0.5}} achieves optimal throughput, maintaining a minimum of 20 FPS across all tested GPUs and reaching peak performance of 28.9 FPS on the H100. 
This performance is enabled by PRQ's optimized configuration (temporal downsampling rate=4, 5 body parts, 4 quantization layers), which requires generating at least 100 tokens per second to sustain 20 FPS operation.
Additional performance optimizations can be achieved by either reducing the number of quantization layers or relaxing part-level control constraints.
Notably, our frame-by-frame decoding strategy proves essential, as conventional layer-by-layer approaches (e.g., RQ) introduce prohibitive delays of tens of seconds in motion generation.

Comparative analysis against MoMask~\cite{guo2024momask} further reveals important architectural trade-offs. 
While both models achieve real-time initial motion generation, \texttt{\textbf{Being-M0.5}} exhibits marginally higher latency due to its expanded token output ($>3\times$) required for part-level control and its larger model architecture. 
However, MoMask's reliance on HM263 features necessitates computationally intensive inverse kinematics (IK) post-processing, whereas our HuMo263 representation enables direct, real-time motion generation with integrated part-level control, completely eliminating the IK computational bottleneck.
This architectural advancement delivers three critical advantages: (1) consistent real-time performance, (2) immediate deployment readiness, and (3) fine-grained part-level controllability.

\subsubsection{Part-level Controllability}
\textbf{\texttt{Being-M0.5}} addresses the fundamental challenge of part-specific motion control through a novel approach that decomposes the human body into five distinct part-level tokens, rather than relying on conventional holistic representations. 
This architectural design enables effective pretraining on fine-grained motion-text pairs and facilitates selective decoding of individual body parts during inference.
While previous works have explored partial body control or pose conditioning in isolation~\cite{zhang2022motiondiffuse,zou2024parco,wang2025fg}, our approach achieves substantially superior controllability through integrated part-aware encoding.
We evaluate \textbf{\texttt{Being-M0.5}} against recent part-specific motion generation methods on the HumanML3D dataset, where it outperforms existing techniques as demonstrated in Table~\ref{tab:control}.
To provide a comprehensive assessment of part-level control precision, we introduce human part awareness as a quantitative controllability metric.
Following the evaluation protocol established by ParCo~\cite{zou2024parco}, we conduct left-right exchange tests on 50 carefully annotated sentences.
Our method achieves a 76\% success rate, significantly surpassing ParCo (64\%) and T2M (46\%), thereby demonstrating significantly more robust and accurate part-aware motion synthesis capabilities.

\input{tables/05_control}

\subsection{Further Discussion}

\subsubsection{How does the VLMM benefit from part-level motion?}
We systematically evaluate the impact of part-level motion representations on the HuMo-I2PM benchmark, with results presented in Table~\ref{tab:part-level}.

First, removing shared joints significantly degrades performance, increasing the FID score from 1.831 to 2.471. 
This deterioration occurs because predicting isolated 3D body parts without joint dependencies presents substantial challenges. 
Given that human joints exhibit highly structured interdependencies where each joint's position is often contingent upon others, sharing joints across different body parts strengthens inter-part connections and consequently improves overall performance.
Second, when comparing PRQ and RQ approaches with part-level descriptions, \texttt{\textbf{Being-M0.5}}-RQ$_6$ underperforms despite utilizing deeper quantization layers and incorporating part labels. 
Additional RQ experiments demonstrate minimal performance gains from part labels alone, thereby underscoring PRQ's effectiveness in learning from fine-grained part-level descriptions. 
Furthermore, \texttt{\textbf{Being-M0.5}}-PRQ$_4$ without part labels performs worse than the corresponding RQ variant, indicating that PRQ's architectural structure and part labels are complementary.
Finally, we investigate the impact of quantization layer depth by increasing PRQ's layer count. 
While deeper layers yield improvements in reconstruction tasks, they instead damage generation performance. 
We attribute this degradation to increased decoding complexity, particularly given \texttt{\textbf{Being-M0.5}}'s frame-by-frame decoding strategy, which contrasts with approaches that generate the base code layer first~\cite{guo2024momask}.

\input{tables/06_part-level}

\subsubsection{Does the visual modality benefit motion pre-training?}
Yes, the results in Table~\ref{tab:abl_vision} demonstrate that \texttt{\textbf{Being-M0.5}} with vision-text-motion alignment consistently outperforms models without visual integration. 
Visual cues provide complementary alignment between visual and textual contexts, offering valuable supervisory signals for motion understanding, particularly when motion data exhibits noise or inconsistencies. 
As a unified multimodal model, VLMM can additionally perform motion estimation to replicate human actions from visual inputs, significantly expanding its application scope. 
We reserve comprehensive exploration of this capability for future work.

\input{tables/07_vision}

\subsubsection{Does the multi-task training increase the controllability of motion generation?}
Yes, multi-task training substantially enhances controllability. 
Table~\ref{tab:multi_task} presents results for different motion task configurations during training, evaluated on the HuMo-T2M testing set. 
The initial data ratio for T2M : I2M : MPI : I2PM : I2LM is configured 5:5:3:2:1.  
Rows 1-3 indicate that removing I2PM or I2LM marginally reduces T2M performance, though the improvements on their respective specialized benchmarks are considerably more significant
Rows 4-5 highlight the critical importance of the Motion Prediction and In-between (MPI) task, reducing the FID score from 6.582 to 6.052 and enabling the VLMM to generate coherent motions from random pose initialization. 
Rows 5-7 confirm the necessity of I2M integration, with the comparison between Row 5 and Row 6 demonstrating that performance improvements stem from task diversity rather than simple data scaling.

\input{tables/08_multitask}

\subsubsection{Motion Tokenizer Selection.}
We analyze the inherent trade-offs in motion tokenizer selection strategies. 
While PRQ adopted by \textbf{\texttt{Being-M0.5}} achieves superior performance on R@\# and MMDist metrics compared to the RQ approach used by Momask, it exhibits higher FID scores on HumanML3D.
We attribute this discrepancy to PRQ's reduced number of quantized layers.
As shown in Row 3 and Row 6 of Table~\ref{tab:part-level}, PRQ exhibits better performance than RQ using the same layers.
Furthermore, Table~\ref{tab:motion_quant} illustrates that FID performance is highly dependent on layer count: with equivalent layer configurations, PRQ achieves competitive reconstruction FID scores.
However, increasing layer depth introduces additional motion tokens, thereby complicating LLM-based decoding processes.
To balance these competing factors, we select PRQ$_4$, which preserves the LLM's advantages in complex tasks such as I2M while maintaining efficient motion representation capabilities.

\begin{figure*}[ht]
\centering   
\begin{subfigure}[b]{1\textwidth}
    \includegraphics[width=\textwidth]{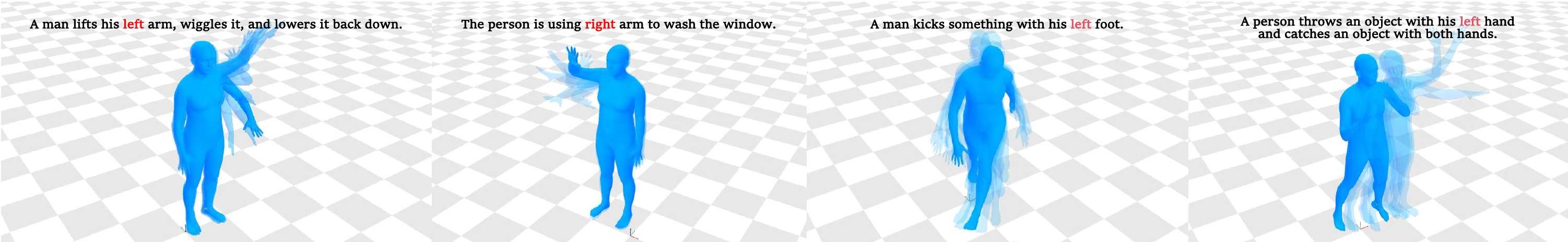}
    \caption{Visualization results of Instruct-to-PartMotion.}
    \label{fig:part_vis}
\end{subfigure}

\vspace{4mm} 

\begin{subfigure}[b]{1.0\textwidth}
    \includegraphics[width=\textwidth]{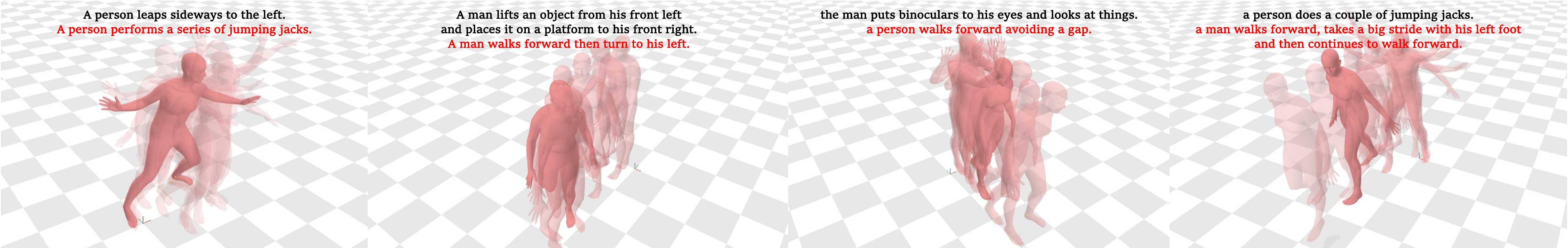}
    \caption{Visualization results of Instruct-to-LongMotion.}
    \label{fig:second}
\end{subfigure}
\caption{Qualitative examples generated by \texttt{\textbf{Being-M0.5}} for Instruct-to-PartMotion (I2PM) and Instruct-to-LongMotion (I2LM). The results demonstrate the ability of our model to generate motion sequences that accurately align with both part-level and long-term instructions.}
\label{fig:visualization}
\end{figure*}

\subsection{Visualization}
As demonstrated in Figure~\ref{fig:visualization}, we present visualization examples showcasing part-level and long-term motion control capabilities generated by \texttt{\textbf{Being-M0.5}}, effectively illustrating the model's fine-grained controllability.
Due to space limitations, additional comprehensive visualizations are provided in Appendix~\ref{app:dataset}.
In contrast to previous models that may require half a minute for motion generation, our approach achieves real-time responsiveness, enabling seamless integration into professional animation workflows with minimal latency overhead, as evidenced in the accompanying figures.

%% file: tables/01_sota_t2m.tex
\begin{table}[ht]
\vspace{0.5cm}
\centering
\setlength{\tabcolsep}{12pt}
\scalebox{0.9}{
\begin{tabular}{lccccc}
\toprule
& LLM & FID $\downarrow$ & R@1 $\uparrow$ & R@3 $\uparrow$ & MMDist  $\downarrow$ \\
\midrule
Real & - & 0.002 & 0.511 & 0.797 & 2.974 \\
\midrule
\rowcolor{BlockA!30}
\multicolumn{6}{l}{\textbf{\# using vanilla VQ-VAE}} \\
\rowcolor{BlockA!30}
MLD~\cite{chen2023executing} & - & 0.473 & 0.481 & 0.772 & 3.196 \\
\rowcolor{BlockA!30}
MotionDiffuse~\cite{zhang2022motiondiffuse} & - & 0.630 & 0.491 & 0.782 & 3.113 \\
\rowcolor{BlockA!30}
T2M-GPT~\cite{zhang2023generating} & GPT-2 & 0.141 & 0.492 & 0.775 & 3.121 \\
\rowcolor{BlockA!30}
MotionGPT$^{1}$~\cite{jiang2023motiongpt} & T5 & 0.162 & 0.409 & 0.667 & 3.992 \\
\rowcolor{BlockA!30}
MotionGPT$^{2}$~\cite{zhang2024motiongpt} & LLaMA-13B  & 0.542 & 0.411 & 0.696& 3.584 \\
\rowcolor{BlockA!30}
MotionLLM~\cite{wu2024motionllm} & Gemma-2b & 0.491 & 0.482 & 0.770 & 3.138 \\
\rowcolor{BlockA!30}
AvatarGPT~\cite{zhou2024avatargpt} & LLaMA-13B  & 0.567 & 0.389 & 0.623& - \\
\rowcolor{BlockA!30}
MotionGPT-v2~\cite{wang2024motiongpt} & LLaMA3-8B & 0.191 & 0.496 & 0.782 & 3.080 \\
\rowcolor{BlockA!30}
LMM~\cite{zhang2024large} & LLaMA3-8B & 0.191 & 0.496 & 0.782 & 3.080 \\
\rowcolor{BlockA!30}
\textbf{Being-M0.5-VQ$_1$} & LLaMA2-7B  & \textbf{0.141} & \textbf{0.528} & \textbf{0.815} & \textbf{2.953} \\
\midrule
\rowcolor{BlockB!30}
\multicolumn{6}{l}{\textbf{\# using advanced VQ variants}} \\
\rowcolor{BlockB!30}
ScaMo-FSQ~\cite{lu2024scamo} & 3B & 0.101 & 0.512 & 0.796  & 2.990 \\
\rowcolor{BlockB!30}
MoMask-RQ$_6$~\cite{guo2024momask} & 760M & \textbf{0.045} & 0.521 & 0.807 & 2.958 \\
\rowcolor{BlockB!30}
\textbf{Being-M0.5-PRQ$_4$} & LLaMA2-7B  & 0.056 & \textbf{0.535} & \textbf{0.821} & \textbf{2.865} \\
\bottomrule
\end{tabular}}
\caption{Comparison with previous motion methods on HumanML3D. Superscripts $^1$ and $^2$ distinguish different works sharing the same model name, while the subscript $n$ in $\mathcal{Q}_n$ denotes the number of quantization layers in the quantizer $\mathcal{Q}$.}

\label{tab:sota_t2m}
\end{table}

%% file: tables/02_sota_i2m.tex
\begin{table}[ht]
\vspace{.3cm}
\centering
\setlength{\tabcolsep}{12pt}
\scalebox{0.9}{
\begin{tabular}{l|c|cccc}
\toprule
& LLM backbone & FID $\downarrow$ & R@1 $\uparrow$ & R@3 $\uparrow$ & MMDist  $\downarrow$ \\
\midrule
T2M-GPT~\cite{zhang2023generating} & GPT-2 & 0.682 & 0.154 & 0.275 & 4.251 \\
MotionGPT$^{1}$~\cite{jiang2023motiongpt} & T5 & 0.382 & 0.268 & 0.352 & 3.621 \\
MotionGPT$^{2}$~\cite{zhang2024motiongpt} & LLaMA-13B  & 0.314 & 0.336 & 0.438  & 3.437 \\
MoMask-RQ$_6$~\cite{guo2024momask} & 760M & 0.324 & 0.325 & 0.382 & 3.441 \\
\midrule
\rowcolor{BlockA!30}
\textbf{Being-M0.5-PRQ$_4$} & LLaMA2-7B  & \textbf{0.148} & \textbf{0.428} & \textbf{0.625} & \textbf{3.259} \\
\bottomrule
\end{tabular}}
\caption{Comparison with prior motion generation methods on the Image-to-Motion (I2M) task using the HuMo-I2M testbed. Note that this evaluation is restricted to approaches with publicly available training code.}
\label{tab:sota_i2m}
\end{table}

%% file: tables/03_unseen.tex
\begin{table}[ht]
\vspace{.3cm}
\centering
\setlength{\tabcolsep}{12pt}
\scalebox{0.9}{
\begin{tabular}{l|c|cccc}
\toprule
Training Data & FID $\downarrow$ & R@1 $\uparrow$ & R@3 $\uparrow$ & MMDist  $\downarrow$ \\
\midrule
HumanML3D~\cite{guo2022generating} & 65.04 & 0.068 & 0.148 & 9.72   \\
MotionX~\cite{lin2024motion} & 43.28 & 0.092 & 0.162 & 8.65  \\
\rowcolor{BlockA!30}
HuMo100M  & \textbf{8.65} & \textbf{0.136} & \textbf{0.245} & \textbf{7.31}  \\
\bottomrule
\end{tabular}}
\caption{Comparison on the Instruct-to-Unseen (I2U) task using the HuMo-Unseen testbed across diverse training datasets.}
\label{tab:unseen}
\vspace{.3cm}
\end{table}

%% file: tables/04_quantization.tex
\begin{table}[ht]
\vspace{.3cm}
\centering
\setlength{\tabcolsep}{12pt}
\renewcommand{\arraystretch}{1}
\scalebox{0.9}{
\small
\begin{tabular}{l|c|cc|cc|cc}
\toprule
\multicolumn{2}{c}{}&\multicolumn{2}{|c}{HumanML3D} &\multicolumn{2}{|c}{Motion-X} & \multicolumn{2}{|c}{HuMo100M}\\
\midrule
Tokenizer & Code & FID & MPJPE  & FID & MPJPE  & FID  & MPJPE \\
\midrule
VQ-VAE$_1$ & 1024  & 0.183 & 47.54 & 0.077 & 38.32 & 5.324 & 123.61\\
H$^2$VQ~\cite{you2022locally} & 512 & - & - & - & 62.34 & - & - \\
RQ-VAE$_6$~\cite{lee2022autoregressive}   & 1024 & 0.032 & 23.58 & 0.035 & 21.64 & 3.928 & 68.17 \\
RQ-VAE$_8$  & 1024 & 0.009 & 20.42 & 0.012 & 18.11 & 3.526 & 64.56 \\
2D-LFQ$_1$~\cite{wang2024quo}  & 16384 & 0.092 & 45.60 & 0.295 & 54.10 & - & - \\
FSQ$_1$  & 65536 & 0.051 & 35.04 & 0.108 & 29.82 & 4.326 & 77.15 \\
\midrule
\rowcolor{BlockA!30}
\textbf{PRQ}$_4$ w/o Shared & 1024 & 0.042 & 19.87 & 0.058 & 23.78 & 3.129 & 48.96 \\
\rowcolor{BlockA!30}
\textbf{PRQ}$_4$ & 1024 & 0.007 & 14.06 & 0.013 & 17.25 & 2.317 & 38.06 \\
\rowcolor{BlockA!30}
\textbf{PRQ}$_6$ & 1024 & \textbf{0.004} & \textbf{13.56} & \textbf{0.007} & \textbf{17.18} & \textbf{2.195} & \textbf{36.47} \\
\bottomrule
\end{tabular}}
\caption{Comparison with previous motion tokenizers. Subscripts in tokenizer names indicate the number of quantization layer, ``w/o Shared'' denotes configuration where different part features exclude shared joints.}
\label{tab:motion_quant}
\end{table}

%% file: tables/05_control.tex
\begin{table}[ht]
\vspace{.3cm}
\centering
\setlength{\tabcolsep}{9pt}
\scalebox{.9}{
\begin{tabular}{l|c|cccc}
\toprule
Method & specialized area & FID $\downarrow$ & R@1 $\uparrow$ & R@3 $\uparrow$ & MMDist  $\downarrow$ \\
\midrule
MotionDiffuse \cite{zhang2024motiondiffuse} & partial control & 0.630 & 0.491 & 0.782 & 3.113 \\
ParCo \cite{zou2024parco} & partial control & 0.109 & 0.515 &0.801 & 2.927 \\
Fg-T2M++\cite{wang2025fg} & fine-grained text control & 0.089 & 0.513 & 0.801 & 2.925 \\
\midrule
\rowcolor{BlockA!30}
Being-M0.5-PRQ$_4$ & - & \textbf{0.056} & \textbf{0.535} & \textbf{0.821} & \textbf{2.865} \\
\bottomrule
\end{tabular}}
\caption{Comparison with previous part-specific text-to-motion methods on HumanML3D.}
\label{tab:control}
\vspace{.3cm}
\end{table}

%% file: tables/06_part-level.tex
\begin{table}[ht!]
\vspace{0.3cm}
\centering
\setlength{\tabcolsep}{12pt}
\scalebox{0.9}{
\begin{tabular}{l|c|c|cccc}
\toprule
 & PT? & FID  & R@1 & R@3  & MMDist  \\
\midrule
Being-M0.5-RQ$_6$ & No & 4.025 & 0.208 & 0.395 & 7.01   \\
Being-M0.5-PRQ$_4$ & No & 4.281 & 0.182 & 0.367 & 7.88   \\
Being-M0.5-RQ$_6$ & Yes & 3.752 & 0.215 & 0.408 & 7.21   \\
Being-M0.5-PRQ$_4$ w/o SHA & Yes & 2.471 & 0.325 & 0.561 & 5.32  \\
Being-M0.5-PRQ$_4$ & Yes & 1.831 & 0.384 & 0.685 & 4.12  \\
Being-M0.5-PRQ$_6$ & Yes & 2.357 & 0.351 & 0.662 & 4.37  \\

\bottomrule
\end{tabular}}
\caption{Comparison of part-level motion configurations on the HuMo-I2PM testbed, where ``PT?'' denotes the inclusion/exclusion of part-level descriptions during training.}
\label{tab:part-level}
\end{table}

%% file: tables/07_vision.tex
\begin{table}[ht!]
\vspace{0.3cm}
\centering
\setlength{\tabcolsep}{12pt}
\scalebox{0.9}{
\begin{tabular}{l|c|cccc}
\toprule
 & FID $\downarrow$ & R@1 $\uparrow$ & R@3 $\uparrow$ & MMDist  $\downarrow$ \\
\midrule
Being-M0.5 w/o 2rd vis & 7.053 & 0.198 & 0.418 & 10.13  \\
Being-M0.5 & 5.791 & 0.206 & 0.445 & 8.85   \\
\bottomrule
\end{tabular}}
\caption{Comparison of text-aligned visual clips on the HuMo-T2M testbed, where ``2nd vis'' indicates the second-stage training for vision-text-motion alignment."}
\label{tab:abl_vision}
\vspace{0.3cm}
\end{table}

%% file: tables/08_multitask.tex
\begin{table}[ht!]
\vspace{.3cm}
\centering
\setlength{\tabcolsep}{15pt}
\scalebox{0.9}{
\begin{tabular}{c|l|c|cccc}
\toprule
& multi-task configuration & FID  & R@1 & R@3  & MMDist  \\
\midrule
1 & T2M+I2M+MPI+I2PM+I2LM & 5.791 & 0.206 & 0.445 & 8.85  \\
2 & T2M+I2M+MPI+I2PM & 5.765 & 0.195 & 0.453 & 9.01   \\
3 & T2M+I2M+MPI+I2LM & 5.952 & 0.192 & 0.453 & 8.91   \\
\midrule
4 & T2M+I2M+MPI & 6.052 & 0.184 & 0.425 & 9.01   \\
5 & T2M+I2M & 6.582 & 0.154 & 0.386 & 9.77   \\
6 & 2$\times$T2M & 7.058 & 0.128 & 0.342 & 11.26   \\
7 & T2M & 7.152 & 0.135 & 0.338 & 10.95   \\
\bottomrule
\end{tabular}}
\caption{Evaluation of multi-task motion training effectiveness on the HuMo-T2M testbed across different configurations.}
\label{tab:multi_task}
\vspace{.3cm}
\end{table}

%% file: sec/6_conclusion.tex
\section{Conclusion}
This paper presents a practical Vision-Language Motion Model (VLMM) for real-time, controllable motion generation that achieves state-of-the-art performance across comprehensive motion benchmarks. 
We develop a systematic data curation pipeline to construct  \texttt{\textbf{HuMo100M}}, the largest motion dataset of its kind containing 100 million instructional instances with part-level descriptions, long-term motion sequences, and temporally aligned visual clips.
Leveraging this extensive dataset, we train \texttt{\textbf{Being-M0.5}}, a unified model that enables fine-grained controllable human motion generation across multiple modalities.
To enhance part-level controllability, we introduce a novel part-aware residual quantization (PRQ) method that serves as an effective motion tokenizer, decomposing human motion into semantically meaningful body-part representations.
Experimental validation demonstrates our model's controllability across diverse benchmarks.
Through systematic ablation studies, we provide key design insights for developing such a practical VLMM.

%% file: sec/7_appendix.tex
In this appendix, we provide comprehensive details and analysis of the \textbf{\texttt{HuMo100M}} dataset in Section~\ref{app:dataset}, followed by an elaborate discussion of the design and implementation of our proposed part-aware residual quantization (PRQ) in Section~\ref{app:prq}.

\section{Additional Details of \textbf{\texttt{HuMo100M}}}
\label{app:dataset}

This section presents detailed information about our large-scale multimodal human motion dataset, \textbf{\texttt{HuMo100M}}.
\textbf{\texttt{HuMo100M}} not only integrates existing publicly available human motion datasets but also substantially expands the scale by extracting numerous motion sequences from web videos using WHAM~\cite{shin2024wham}.

\begin{figure*}[ht]
\vspace{.3cm}
  \centering
\includegraphics[width=.95\textwidth]{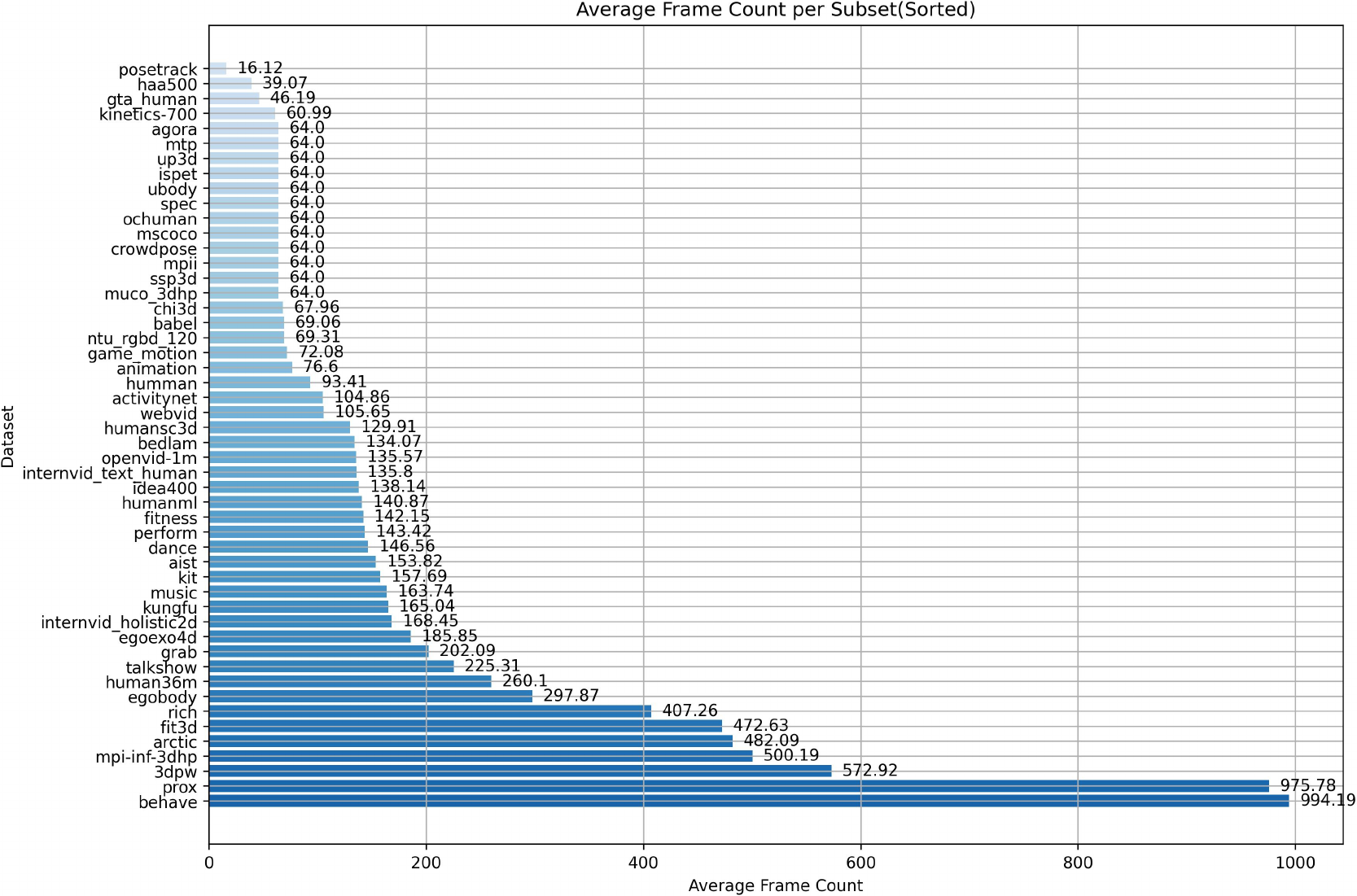}
  \caption{Distribution of motion sequence lengths across different subsets in \textbf{\texttt{HuMo100M}}}
  \label{fig:app_frame_count}
\vspace{.3cm}
\end{figure*}

\subsection{Statistic Analysis of Data and Word Distribution}
\begin{figure*}[ht]
  \centering
\includegraphics[width=.95\textwidth]{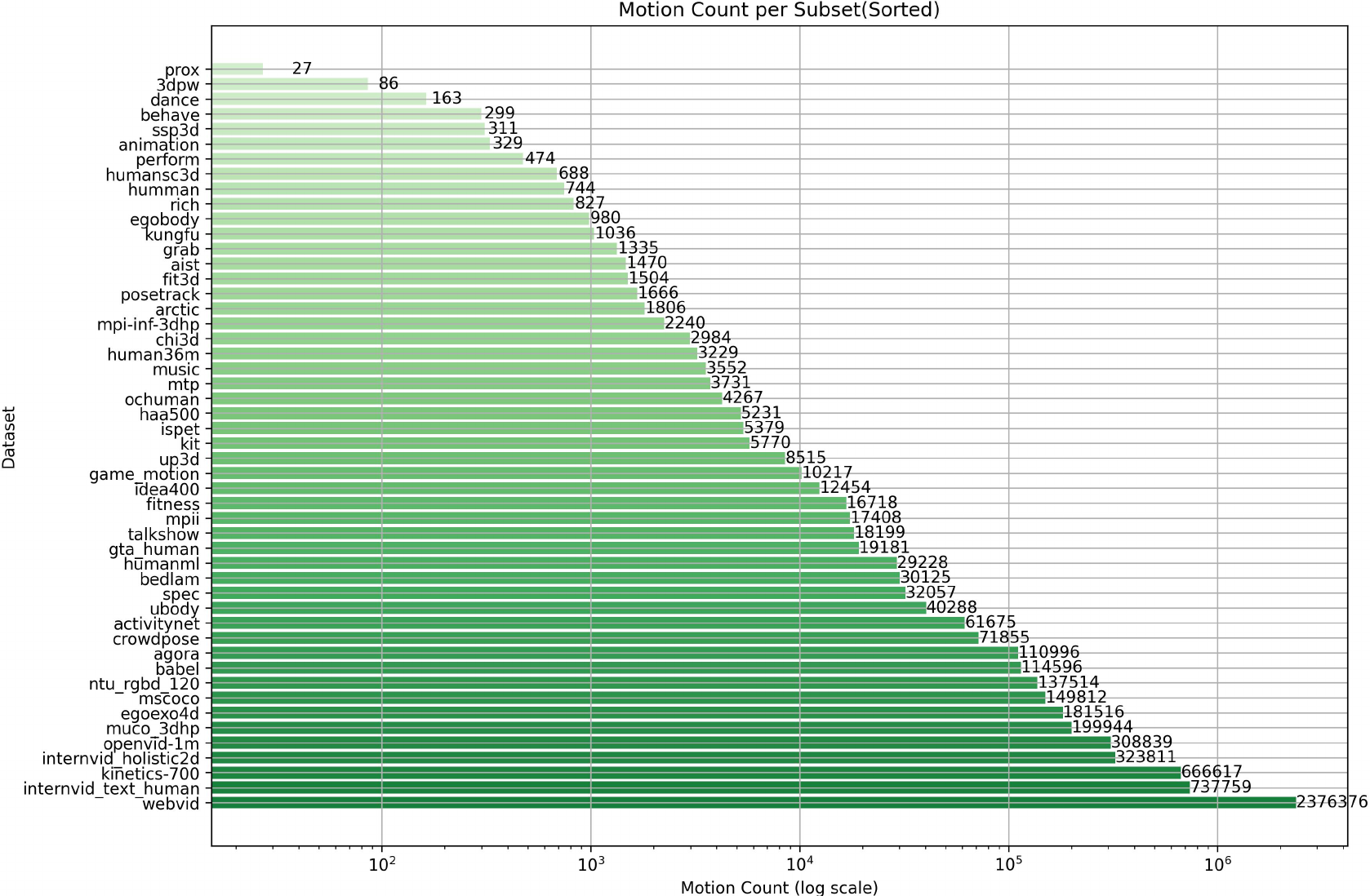}
  \caption{Distribution of motion instance counts across different subsets in \textbf{\texttt{HuMo100M}} (logarithmic scale)}
  \label{fig:app_motion_count}
\end{figure*}

\noindent\textbf{Data Distribution.}
Figure~\ref{fig:app_frame_count} illustrates the distribution of motion sequence lengths across different subsets of \textbf{\texttt{HuMo100M}}. 
The analysis reveals that \textbf{\texttt{HuMo100M}} successfully integrates motion sequences from established datasets, including Motion-X~\cite{lin2023motion}, 3DPW~\cite{von2018recovering}, and MSCOCO~\cite{lin2014microsoft}.
Notably, PoseTrack~\cite{andriluka2018posetrack} contains the shortest average sequence length (16.12 frames), indicating a focus on brief motion segments. In contrast, PROX~\cite{hassan2019resolving} and BEHAVE~\cite{bhatnagar2022behave} feature substantially longer sequences (994.19 and 975.78 frames, respectively), demonstrating that the dataset encompasses diverse motion durations.
Figure~\ref{fig:app_motion_count} presents the distribution of motion instance counts across different \textbf{\texttt{HuMo100M}} subsets on a logarithmic scale, highlighting significant variations in dataset sizes. 
The number of motion instances ranges from 27 sequences in PROX~\cite{hassan2019resolving} to 2,376,376 motion sequences in WebVid~\cite{Bain21}, illustrating the comprehensive scale and diversity of our compiled dataset.

Figure \ref{fig:app_motion_count} shows the distribution of the number of motions in different subsets of \textbf{\texttt{HuMo100M}} (logarithmic scale) and demonstrates the difference in the number of motion instances across subsets, ranging from 27 motions in PROX\cite{hassan2019resolving} to 2,376,376 motions in Webvid\cite{Bain21}.

\begin{figure*}[ht]
\centering
\includegraphics[width=0.9\linewidth]{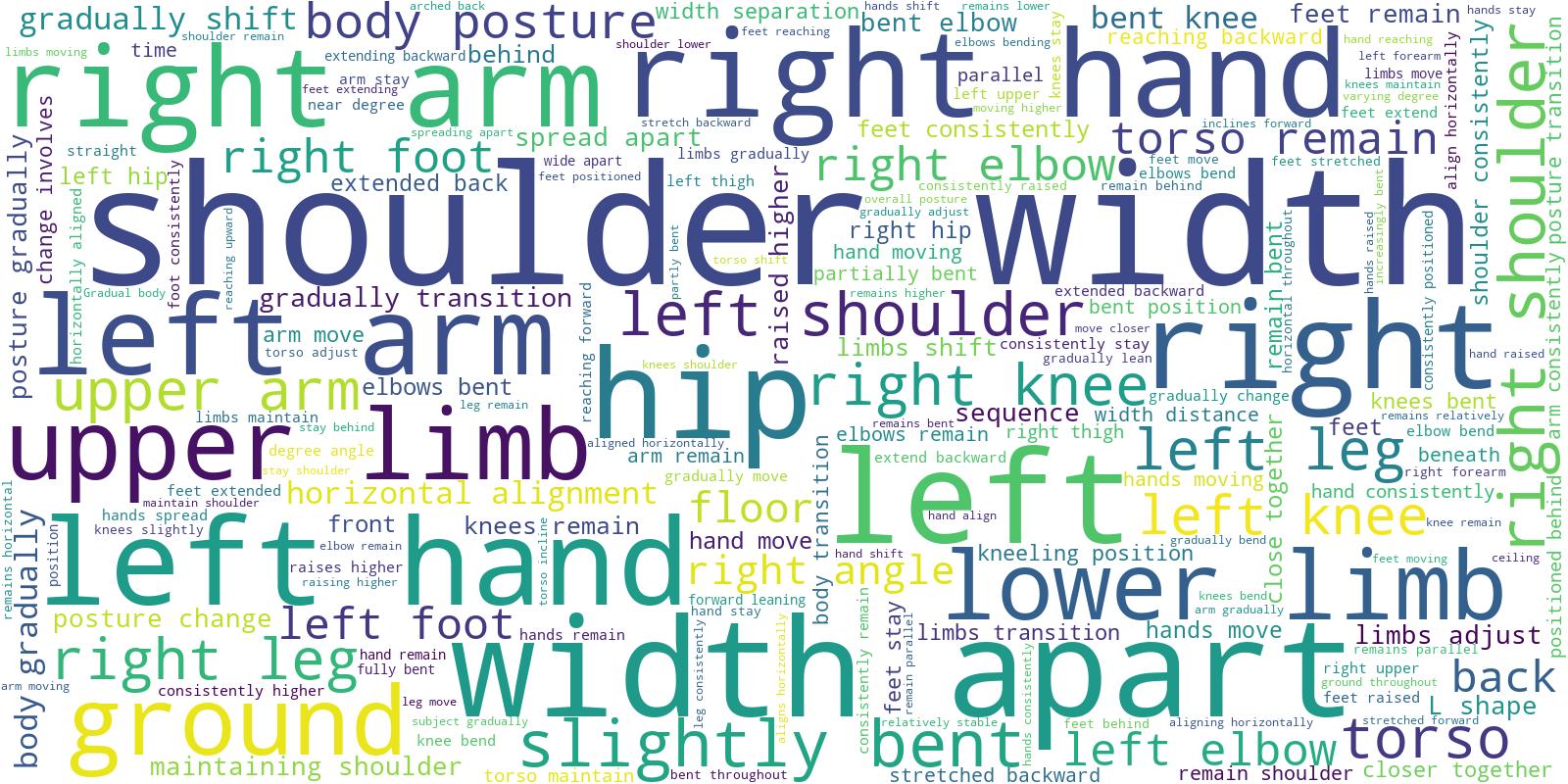}
\caption{Word cloud visualization of rule-based motion descriptions in \textbf{\texttt{HuMo100M}}, highlighting semantic relationships and positional constraints between joints.}
\label{fig:wordcloud_text_rule}
\end{figure*}

\begin{figure*}[ht]
\centering
\includegraphics[width=0.9\linewidth]{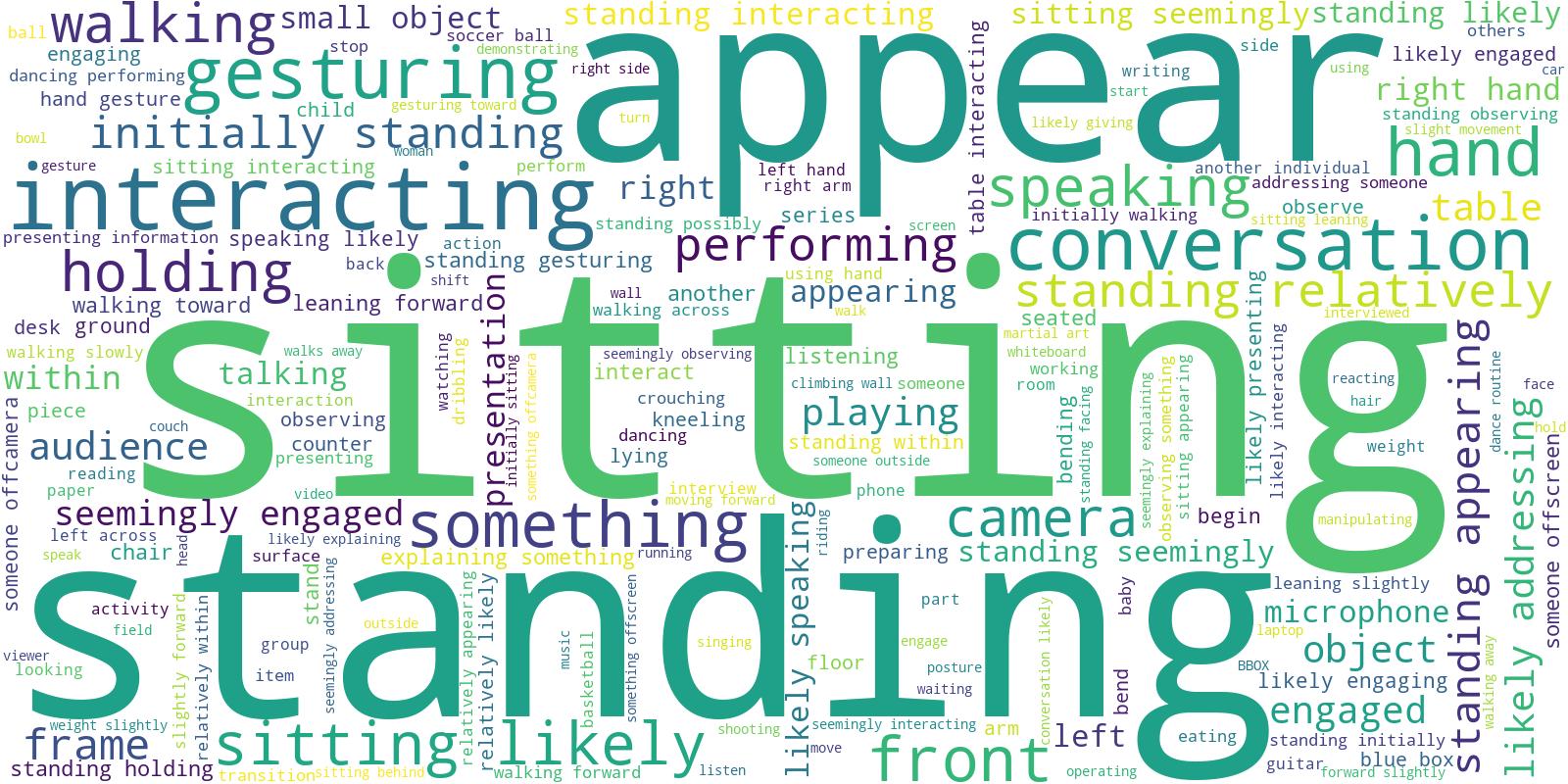}
\caption{Word cloud visualization of body-level motion descriptions in \textbf{\texttt{HuMo100M}}, emphasizing high-level human activities and movements.}
\label{fig:wordcloud_text_body}
\end{figure*}

\begin{figure*}[ht]
\centering
\includegraphics[width=0.9\linewidth]{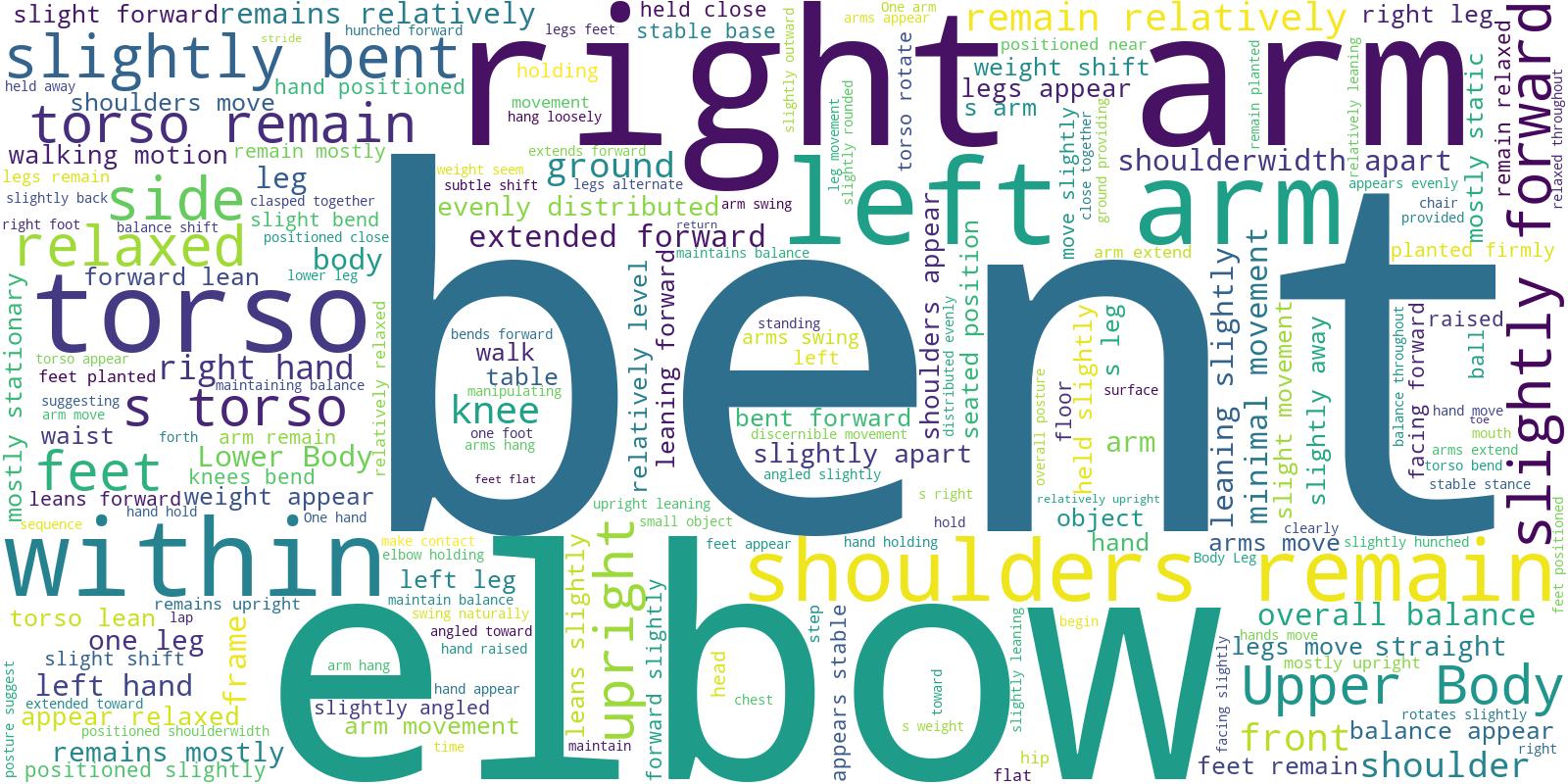}
\caption{Word cloud visualization of part-level motion descriptions in \textbf{\texttt{HuMo100M}}, focusing on detailed movements of specific body regions.}
\label{fig:wordcloud_text_part}
\end{figure*}

\noindent\textbf{Word Distribution.}
To investigate the linguistic characteristics of annotated motion text within \textbf{\texttt{HuMo100M}}, we generate word clouds from the complete text corpus to visualize distinct linguistic patterns. 
We compute three separate word clouds corresponding to rule-based, part-level, and body-level text annotations, respectively. 
Figure~\ref{fig:wordcloud_text_rule} demonstrates that rule-based descriptions emphasize semantic relationships between joints and positional constraints within the kinematic structure. 
In contrast, Figure~\ref{fig:wordcloud_text_part} illustrates that part-level descriptions focus on detailed movements of specific anatomical regions, including the torso, shoulders, legs, and arms. Figure~\ref{fig:wordcloud_text_body} reveals that body-level annotations predominantly capture high-level human activities such as standing, sitting, and walking motions. 
This hierarchical annotation structure, spanning rule-based, part-level, and body-level descriptions, provides comprehensive semantic coverage that facilitates enhanced multimodal alignment in VLMMs across different levels of motion granularity.

\begin{figure*}[ht!]
\centering
\includegraphics[width=0.9\textwidth]{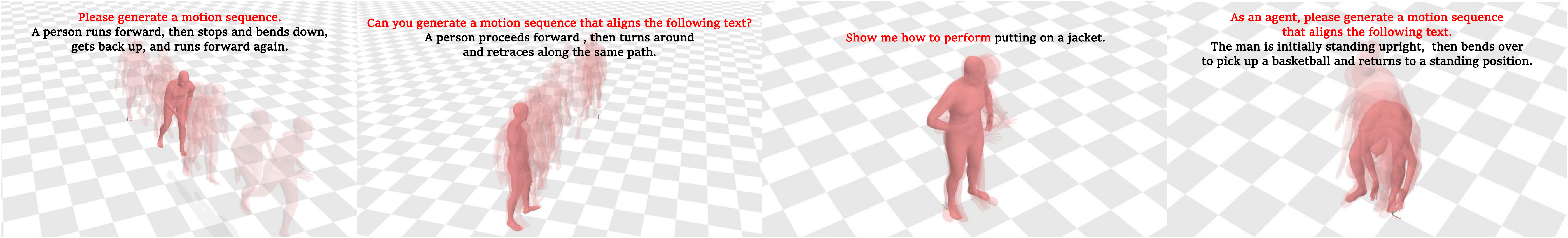}
\caption{Visualization results generated by \textbf{\texttt{Being-M0.5}} from diverse natural language instructions, demonstrating the model's versatility in interpreting and executing varied motion commands.}
\label{fig:app_vis_ric}
\end{figure*}

\begin{figure*}[ht!]
\centering
\includegraphics[width=0.9\textwidth]{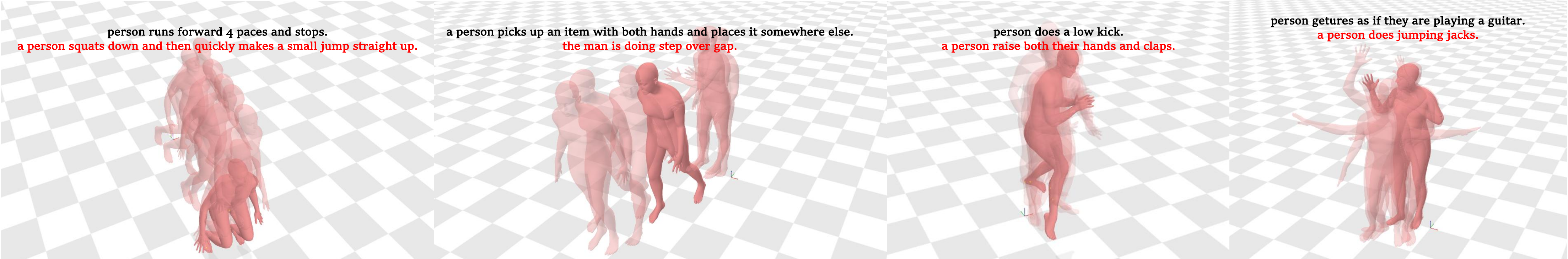}
\caption{Visualization results generated by \textbf{\texttt{Being-M0.5}} given the long-term instruction showcasing the model's capability for coherent long-duration motion synthesis.}
\label{fig:app_vis_long}
\end{figure*}

\begin{figure*}[ht!]
\centering
\includegraphics[width=0.9\textwidth]{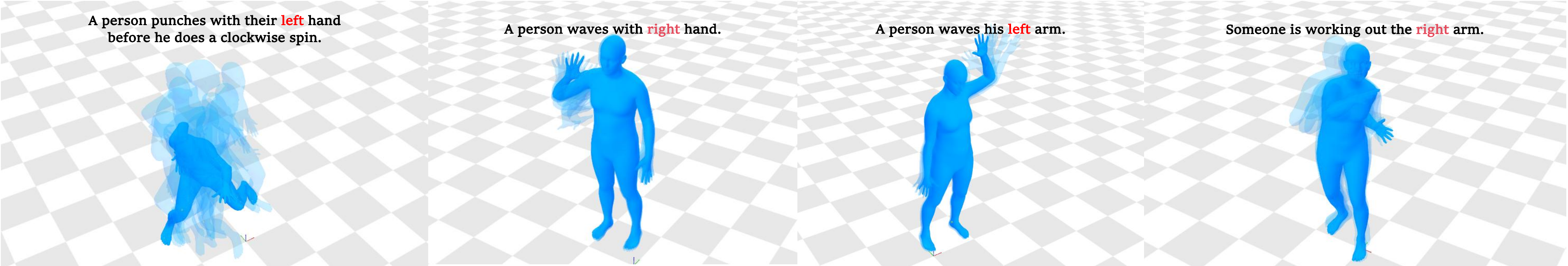}                                                               
\caption{Visualization results generated by \textbf{\texttt{Being-M0.5}} givne the part-level instruction, illustrating fine-grained controllability over individual anatomical regions.}
\label{fig:app_vis_part}
\end{figure*}

\begin{figure}[ht]
\centering
\includegraphics[width=0.6\textwidth]{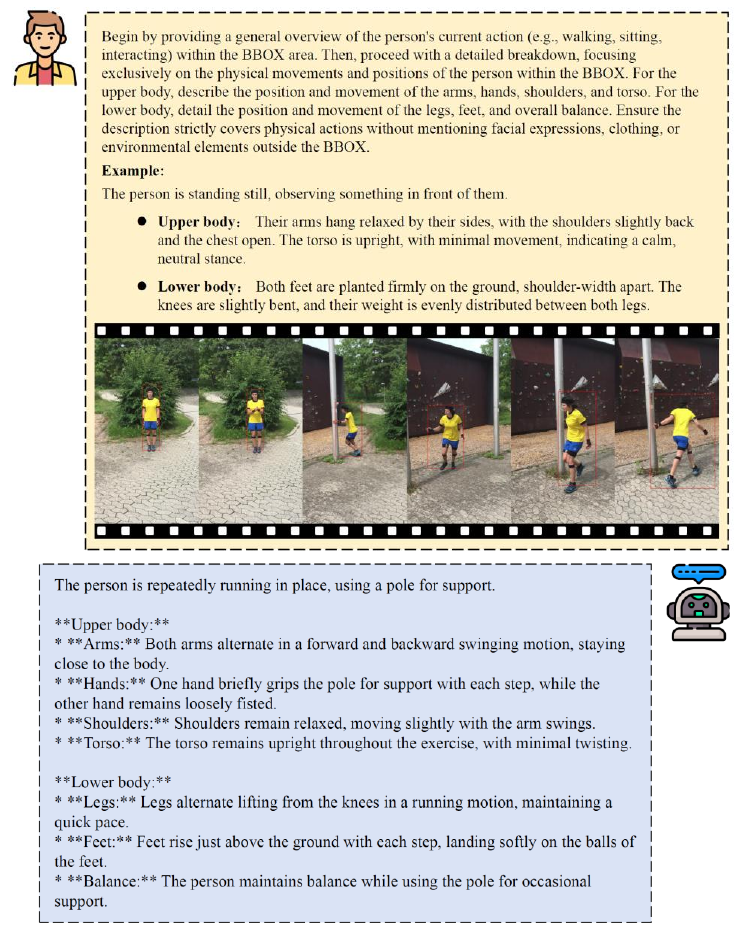}
\caption{Prompt template utilized for generating part-level motion descriptions from video content using large multimodal models (LMMs) such as Gemini-1.5-Pro and GPT-4o-mini. Each sample in \textbf{\texttt{HuMo100M}} includes both ``body-level'' (top) and ``part-level'' (bottom) annotations to distinguish between whole-body and partial motion descriptions.}
\label{fig:app_prompt_template}
\end{figure}

\subsection{Instruction Generation}
To fully exploit the potential of the \textbf{\texttt{HuMo100M}} dataset and support diverse downstream applications (including part-level motion control and vision-based motion understanding), we generate comprehensive, high-quality instructional annotations.
These annotations encompass not only traditional whole-body motion descriptions but also incorporate fine-grained descriptions targeting individual anatomical regions.
We primarily generate these instructions in two ways:

\begin{itemize}[leftmargin=1.5em]
\item \textbf{LMM-based Generation:} 
We design comprehensive prompt templates (illustrated in Figure~\ref{fig:app_prompt_template}) and leverage powerful LMMs, including Gemini-1.5-Pro~\cite{team2024gemini}, to extract detailed part-level motion descriptions from video sequences.
This approach enables the generation of semantically rich, natural language descriptions that capture nuanced motion characteristics.    
\item \textbf{Rule-based Generation (PoseScript):} 
Complementing LMM-generated descriptions, we utilize PoseScript~\cite{delmas2024posescript} and \textit{posecodes} to extract structured semantic pose information. 
This methodology enables the generation of precise geometric instructions describing spatial relationships between joints, such as ``the left hand is positioned below the right hand''.
\end{itemize}

\subsection{Visualization Examples}

Figures~\ref{fig:app_vis_ric}, \ref{fig:app_vis_long}, and \ref{fig:app_vis_part} present comprehensive motion visualization results generated by \textbf{\texttt{Being-M0.5}} across diverse instruction modalities, including random natural language commands, initial pose conditioning, long-term sequential instructions, and part-level control directives. 
These visualizations demonstrate \textbf{\texttt{Being-M0.5}}'s robust capability to interpret and execute instructions across varied formats, showcasing exceptional versatility and reliability in motion generation tasks across different complexity levels and temporal scales.

\section{Part-level Residual Quantization}
\label{app:prq}

In our Part-level Residual Quantization (PRQ) approach, we decompose the complete human skeletal structure into five anatomically meaningful regions:

\begin{itemize}[leftmargin=1.5em]
\item \textbf{Left Hand}: spine$_1$, spine$_2$, spine$_3$, left collar, left shoulder, left elbow, left wrist
\item \textbf{Right Hand}: spine$_1$, spine$_2$, spine$_3$, right collar, right shoulder, right elbow, right wrist
\item \textbf{Left Leg}: spine$_1$, spine$_2$, spine$_3$, left hip, left knee, left ankle, left foot
\item \textbf{Right Leg}: spine$_1$, spine$_2$, spine$_3$, right hip, right knee, right ankle, right foot
\item \textbf{Torso}: spine$_1$, spine$_2$, spine$_3$, neck, left collar, right collar, head
\end{itemize}

The core structural elements --- pelvis, spine$_1$, spine$_2$, spine$_3$ --- are shared across all body parts, as they provide fundamental kinematic stability during human motion. 
Each joint is represented using relative 6D rotations and complementary 3D positional information, resulting in 63+8 dimensional features per body part, which includes 4D root node transformation and 4D foot contact state information. When aggregating part-specific features into unified motion features, we compute the average of shared joint features to maintain kinematic consistency.